\documentclass{SAI}

\usepackage{natbib}
\usepackage{xspace}
\usepackage{graphicx}
\usepackage{multirow}

\usepackage{amsmath,amssymb,amsfonts}
\usepackage{mathrsfs}
\usepackage[title]{appendix}
\usepackage{textcomp}
\usepackage{manyfoot}
\usepackage{booktabs}
\usepackage{makecell}
\usepackage{threeparttable}
\usepackage{caption}
\usepackage{array}
\usepackage{algorithm}
\usepackage{algorithmicx}
\usepackage{algpseudocode}
\usepackage{listings}
\usepackage{float}
\usepackage[section]{placeins}
\usepackage{ragged2e}
\usepackage{enumitem}
\usepackage[most]{tcolorbox}
\usepackage{xltabular}
\usepackage[colorlinks=true,linkcolor=SAIgreen,citecolor=SAIgreen,urlcolor=Scimaster_Blue]{hyperref}

\definecolor{BrickRed}{RGB}{178,34,34}
\definecolor{OliveGreen}{RGB}{85,107,47}

\lstdefinelanguage{yaml}{
  keywords={true,false,null,y,n,yes,no,on,off},
  sensitive=false,
  comment=[l]{\#},
  morestring=[b]",
  morestring=[b]'
}

\lstdefinelanguage{markdown}{
  sensitive=false,
  comment=[l]{\#},
  morestring=[b]",
  morestring=[b]'
}

\lstdefinestyle{RoboCoachCode}{
  basicstyle=\ttfamily\footnotesize,
  columns=fullflexible,
  keepspaces=true,
  breaklines=true,
  breakatwhitespace=false,
  breakautoindent=true,
  breakindent=0pt,
  postbreak=\mbox{\textcolor{gray}{$\hookrightarrow$}\space},
  showstringspaces=false,
  tabsize=2
}

\lstdefinestyle{RoboCoachMarkdown}{
  style=RoboCoachCode,
  language=markdown
}

\newtcolorbox{PRDMarkdownBox}[1][]{
  enhanced,
  breakable,
  colback=gray!5,
  colframe=gray!50,
  boxrule=0.5pt,
  arc=2pt,
  left=6pt,
  right=6pt,
  top=6pt,
  bottom=6pt,
  #1
}

\newcommand{\PRDListingOrPlaceholder}[1]{
  \IfFileExists{#1}{
    \lstinputlisting[style=RoboCoachMarkdown]{#1}
  }{
    \texttt{PRD source missing: \detokenize{#1}}
  }
}

\newcommand{\NA}{N/A}
\newcommand{\ExtendedDataTableShift}{-1.5cm}
\title{From Digital to Physical: Digital Agents as Autonomous Coaches for Physical Intelligence}

\author{Zixing Lei}
\affiliation{School of Artificial Intelligence, Shanghai Jiao Tong University, Shanghai, China}
\affiliation{Zhongguancun Academy, Beijing, China}
\equal

\author{Genjia Liu}
\affiliation{School of Integrated Circuits, Shanghai Jiao Tong University, Shanghai, China}
\equal

\author{Yuanshuo Zhang}
\affiliation{School of Computer Science, Shanghai Jiao Tong University, Shanghai, China}
\equal

\author{Qipeng Liu}
\affiliation{School of Artificial Intelligence, Shanghai Jiao Tong University, Shanghai, China}

\author{Yuzhu Cai}
\affiliation{School of Artificial Intelligence, Shanghai Jiao Tong University, Shanghai, China}

\author{Sixiang Chen}
\affiliation{State Key Laboratory of Multimedia Information Processing, School of Computer Science, Peking University, Beijing, China}

\author{Jixian Wu}
\affiliation{State Key Laboratory of Multimedia Information Processing, School of Computer Science, Peking University, Beijing, China}

\author{Yunhong Wang}
\affiliation{Zhongguancun Academy, Beijing, China}

\author{Weixin Li}
\affiliation{Zhongguancun Academy, Beijing, China}

\author{Chuan Wen}
\affiliation{School of Artificial Intelligence, Shanghai Jiao Tong University, Shanghai, China}

\author{Bo Zhao}
\affiliation{School of Artificial Intelligence, Shanghai Jiao Tong University, Shanghai, China}

\author{Shanghang Zhang}
\affiliation{State Key Laboratory of Multimedia Information Processing, School of Computer Science, Peking University, Beijing, China}

\author{Wenzhao Lian}
\affiliation{School of Artificial Intelligence, Shanghai Jiao Tong University, Shanghai, China}

\author{Siheng Chen}
\affiliation{School of Artificial Intelligence, Shanghai Jiao Tong University, Shanghai, China}

\begin{document}
\maketitle

\begin{center}
\small Correspondence: chezacarss@sjtu.edu.cn; sihengc@sjtu.edu.cn
\end{center}

\begin{abstract}
Digital agents built on large language models can now solve complex coding and scientific tasks, but it remains unclear whether such digital competence can be converted into physical competence. We study autonomous embodied policy development: whether an agent, given only a natural-language task specification, can implement, train, diagnose and revise robot policies using physically grounded feedback.
    We introduce \textsc{RoboCoach}, an autonomous multimodal agent system, and \textsc{RoboCoach-Bench}, a 32-task benchmark spanning four simulation platforms with expert-authored references, complemented by real-robot transfer experiments.
    Across seven frontier language models, \textsc{RoboCoach} matches and, in aggregate, exceeds platform references under binary task-completion criteria.
    Ablations show that the gains come from the closed-loop framework--execution feedback, training signals, rollout-video diagnostics, memory and search--not only from model strength.
    Policies optimised by \textsc{RoboCoach} also transfer to physical robots in two hardware laboratories, demonstrating a bounded route from digital agents to physical competence.
\end{abstract}

\noindent\textbf{Keywords:} embodied AI, long-horizon autonomy, large language models, reinforcement learning, imitation learning

\newpage

% Introduction begins here (Nature Machine Intelligence style: no heading).
    \begin{figure}[!t]
        \centering
        \begin{minipage}[t]{0.95\textwidth}
            \centering
            \parbox[c]{2.2em}{\raggedright\textbf{\large a}}%
            \parbox[c]{\dimexpr\linewidth-2.2em\relax}{\centering\includegraphics[width=0.96\linewidth]{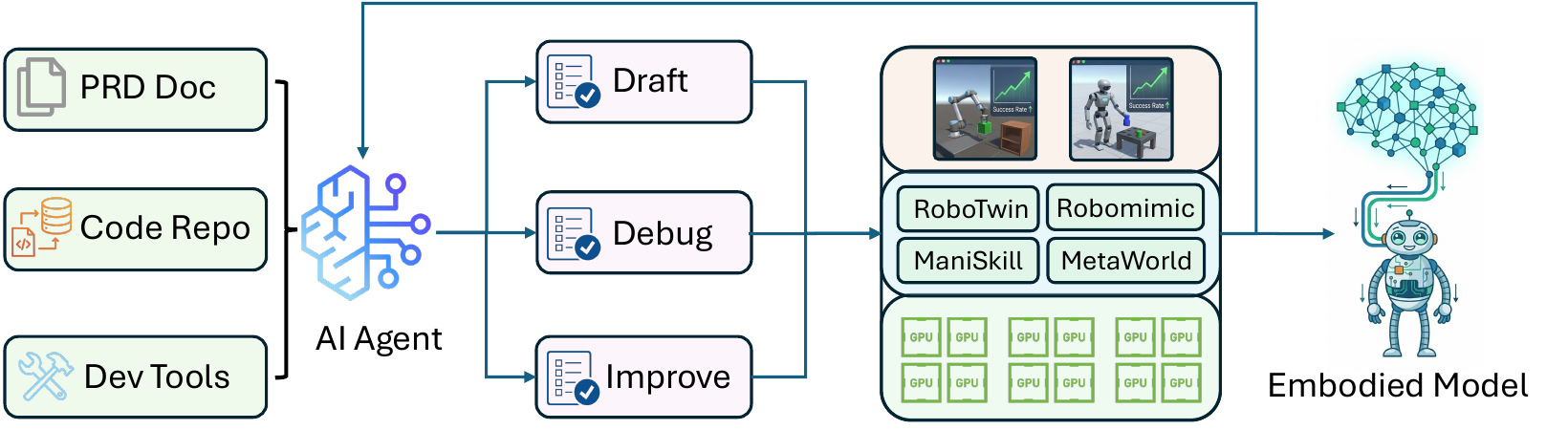}}
        \end{minipage}
    
        \vspace{0.9em}
    
        \begin{minipage}[t]{0.95\textwidth}
            \centering
            \parbox[c]{2.2em}{\raggedright\textbf{\large b}}%
            \parbox[c]{\dimexpr\linewidth-2.2em\relax}{\centering\includegraphics[width=0.96\linewidth]{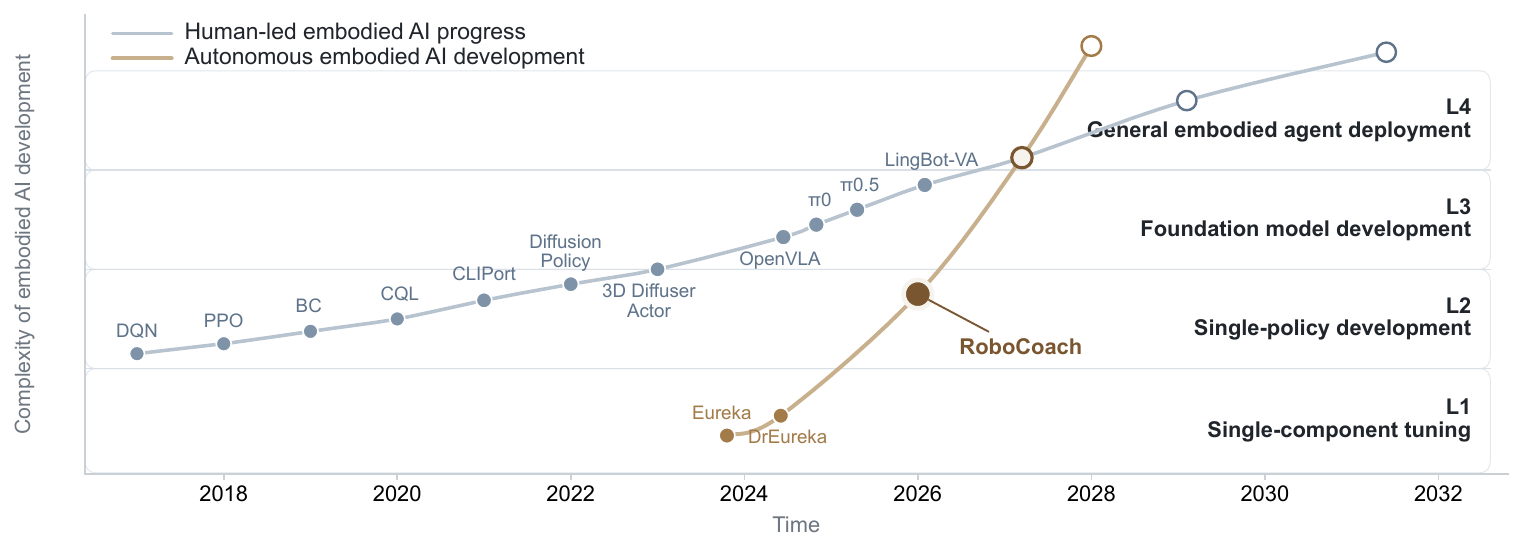}}
        \end{minipage}
        \captionsetup{justification=RaggedRight, singlelinecheck=false}
        \caption{\textbf{Core idea and capability roadmap of \textsc{RoboCoach}.}
        \textbf{a,} \textsc{RoboCoach} is a fully autonomous LLM agent system that aligns with human objectives and iteratively evolves embodied policies through feedback-driven optimisation.
        \textbf{b,} Roadmap of autonomous embodied AI development. The blue trajectory denotes historical human-led progress in embodied AI, whereas the gold trajectory denotes the emerging autonomous trajectory in which AI systems increasingly participate in developing and improving robotic capabilities. The vertical axis indicates increasing complexity of embodied AI development: Level~1 corresponds to optimising individual components within a closed training pipeline, such as reward functions or domain-randomisation schemes; Level~2 corresponds to developing or improving a single policy in a closed benchmark or task setting, where we position \textsc{RoboCoach}; Level~3 corresponds to developing embodied foundation models across embodiments, tasks, and more open environments; and Level~4 corresponds to developing, deploying, and maintaining general embodied agent systems in open-world operation. Higher levels indicate broader development scope and system complexity, rather than fully demonstrated capabilities of current systems.}
        \label{fig:cover}
    \end{figure}
    Recent progress in LLM foundation models has substantially improved long-horizon reasoning, planning, and code generation~\cite{qwen3tech,glm45,Guo_2025}.
    Built on these advances, agentic systems that couple reasoning, tool use, and verbal self-reflection now autonomously complete a broad range of digital tasks~\cite{react,shinn2023reflexion}.
    In mathematics and theoretical physics, they have begun to support theorem proving, symbolic derivation, and algorithmic discovery; in machine learning engineering, they now assist with repository-scale code editing, experiment scripting, debugging, and evaluation orchestration across complex training pipelines, and have begun to close end-to-end loops of scientific hypothesis generation and validation~\cite{AlphaGeometry,alphaevolve,codex,swebench,mlebench,gottweis2025aicoscientist}.
    Recent CLAW-style frameworks further suggest that agents can create new skills and tools for themselves, expanding their own digital capabilities~\cite{openclaw,clawe}.
    Given these results within digital space, a sharper question naturally follows: can a fully autonomous digital agent also create the tools it needs for the physical world, and thereby produce embodied intelligence?
    
    This question carries weight beyond engineering.
    From a conceptual standpoint, it probes whether intelligence developed entirely in digital space can produce competence in the physical world, a possibility that decades of philosophical debate have treated with scepticism.\footnote{We refer to classical arguments about symbolic grounding, Chinese-room reasoning, behaviour-based robotics and physical symbol systems as conceptual motivation, not as a claim to resolve the debate~\cite{harnad1990,searle1980,brooks1991,newell1976}.}
    In this work we address the operational level directly: we study autonomous embodied policy development, asking whether a digital agent, given only a natural-language task specification, can carry out the full iterative loop of a robotics researcher, from implementation and training to behavioural diagnosis and revision from physically grounded feedback.
    A positive answer would also inform the conceptual debate by showing that a purely digital agent can produce physically competent behaviour through closed-loop interaction with embodied environments.
    
    A roadmap of autonomous embodied AI development (Fig.~\ref{fig:cover}b) makes this asymmetry explicit.
    Human-led embodied AI has progressed from task-specific policy engineering toward broader embodied foundation models that span multiple embodiments, tasks, datasets, and deployment conditions~\cite{pi0,intelligence2025pi05,pi06,liu2024rdt,generalist2026gen1}.
    Digital agents have also advanced rapidly in purely digital domains, where they can now participate in coding, debugging, evaluation, and parts of their own development loops.
    By contrast, autonomous agents for embodied AI remain much closer to component-level automation: existing systems can generate rewards, propose tasks, synthesize environments, or tune isolated modules, but they do not yet close the loop from a task specification to a physically competent policy~\cite{Eureka,text2reward,DrEureka,codeaspolicy,RoboGen,gensim}.
    
    This lag reflects an \emph{embodied competence gap}.
    Embodied policy development produces rich feedback signals, including execution errors, training curves, scalar task success, rollout videos, behavioural traces, and failure modes.
    Yet there is no autonomous system that systematically uses these signals to diagnose embodied failure and revise the resulting policy-development process.
    The gap is therefore not simply between writing code and running training, but between producing digital artefacts and producing policies that succeed under a physical task-completion criterion.
    It appears in two forms.
    The first is methodological: autonomous agents lack a closed-loop mechanism for converting textual, quantitative, and perceptual feedback from embodied rollouts into policy revision.
    The second is evaluative: existing embodied benchmarks are primarily designed to compare policies, learning algorithms, or robot systems, not autonomous agents that produce policies from task specifications.
    Evaluating such agents requires a benchmark that exposes the full policy-development loop, pairs each task with expert-authored baselines, and applies a common physical task-completion metric across heterogeneous platforms and policy families.

    To overcome the methodological obstacle, we present \textsc{RoboCoach}, an autonomous agent system designed to close the embodied policy-development loop within a single framework.
    For coding and training, the agent uses textual execution signals and quantitative training signals to detect implementation and optimisation failures.
    For behavioural diagnosis, which cannot be resolved from symbolic and scalar feedback alone, a vision-language model converts rollout videos into structured diagnostics of the robot's physical behaviour, giving the agent perceptual evidence that a human researcher would also use to guide revision.
    Across iterations, a tree-structured memory preserves each complete physical experiment, and a feedback agent uses this history to decide whether to continue, revert, or restart.

    \begin{figure}[!t]
        \centering
        \begin{minipage}[t]{0.99\textwidth}
            \centering
            \includegraphics[width=0.97\textwidth]{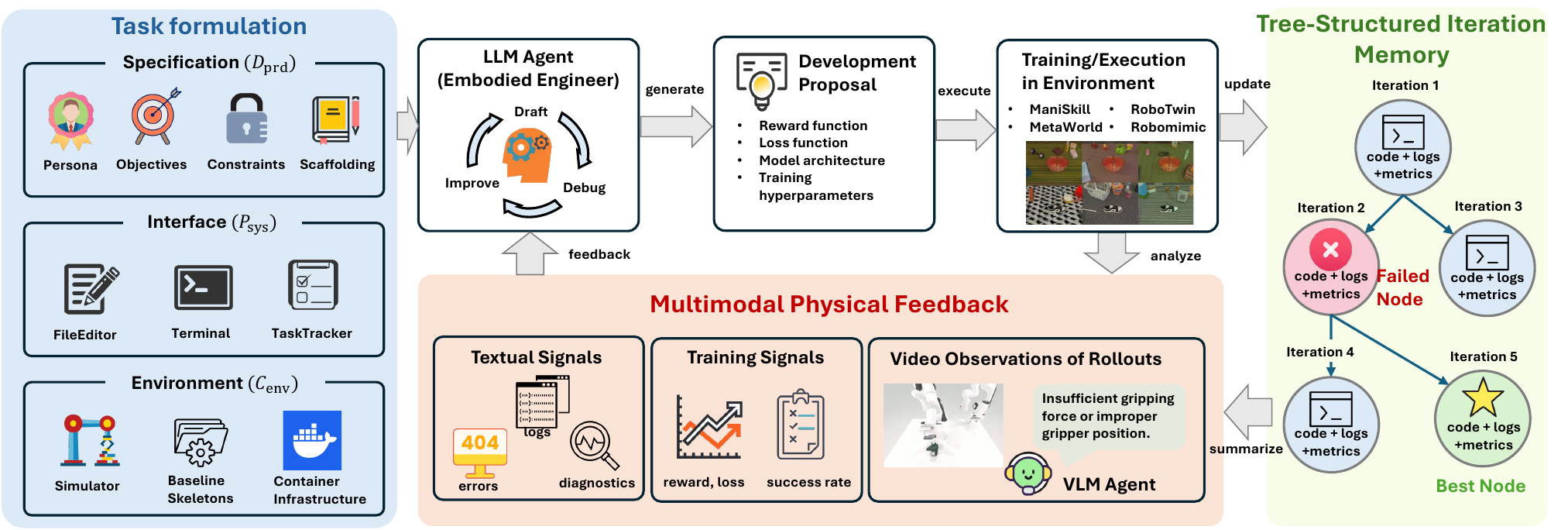}
        \end{minipage}
        \captionsetup{justification=RaggedRight, singlelinecheck=false}
        \caption{\textbf{\textsc{RoboCoach} closes the embodied policy-development loop.}
        An LLM-based agent turns a natural-language task specification into executable training proposals, uses execution logs, training signals and rollout-video diagnostics to revise policy-development code, and stores evaluated attempts in a tree-structured memory for feedback-guided branching.}
        \label{fig:framework_and_benchmark}
    \end{figure}
    
    To address the evaluation obstacle, we introduce \textsc{RoboCoach-Bench}, a benchmark designed to evaluate autonomous embodied policy development rather than only the policies or algorithms that result from it.
    Each task gives an agent a natural-language task specification and evaluates whether the agentic loop can produce a policy that completes the task.
    The benchmark spans 32 tasks across four simulation platforms (ManiSkill~\cite{maniskill3}, RoboTwin~\cite{robotwin2}, Robomimic~\cite{robomimic}, MetaWorld~\cite{metaworld}), covers both reinforcement learning and imitation learning, and includes policy architectures from MLPs and action chunking transformers~\cite{zhao2023ACT} to diffusion policies~\cite{chi2025DP} and vision-language-action models~\cite{rt-2,openvla,pi0,intelligence2025pi05}.
    This heterogeneity tests performance across different development settings rather than a single codebase or policy family.
    Each task is paired with an expert-authored baseline from the corresponding embodied-AI platform, enabling direct comparison between agent-produced and human-engineered policy-development pipelines.
    Performance is measured by physical task completion rate over 100 episodes, with real-robot transfer experiments used as a complementary test of whether simulation-derived improvements extend beyond the benchmark.
    
    Experiments across seven frontier models yield three findings.
    First, across \textsc{RoboCoach-Bench}, \textsc{RoboCoach} paired with strong LLMs matches and, in aggregate, exceeds expert-authored baselines provided by the leading research groups behind these widely used embodied-AI platforms, demonstrating that an autonomous digital agent can compete with human-engineered embodied policy-development pipelines under a physically grounded success criterion.
    Second, these gains arise not only from the underlying large language models, but also from the \textsc{RoboCoach} framework, which combines physically grounded feedback, memory, and search. For example, pairing Gemini 3.0 Pro and GPT-5.2 with \textsc{RoboCoach} improves their benchmark-wide performance by 40.5 and 36.4 percentage points, respectively.
    Third, the benchmark reveals a systematic gap between producing runnable or trainable solutions and achieving physical task success, which \textsc{RoboCoach} substantially reduces. We pair this simulation-based evidence with real-robot transfer on multiple tasks across two hardware laboratories, in which policies optimised by \textsc{RoboCoach} transfer successfully to physical robots.
    
    Returning to the question posed at the outset: can digital intelligence produce physical competence? Within our experimental scope, the answer is yes: \textsc{RoboCoach} substantially closes the embodied competence gap, with gains matching or exceeding human experts in simulation and transferring to physical robots. Conceptually, this provides empirical support for the view that digitally developed intelligence can produce physically competent behaviour when coupled to embodied feedback and evaluation. Practically, it points toward a new mode of embodied-AI development in which the routine cycle of implementation, training, failure diagnosis, and revision is delegated to digital agents, substantially reducing the per-task human engineering effort that currently bottlenecks the field.

    \section{Results}\label{sec:results}

    \subsection{Experimental setting}\label{subsec:setup}
    We evaluate \textsc{RoboCoach} with \textsc{RoboCoach-Bench}, an end-to-end benchmark designed to test whether a digital agent can convert a natural-language task specification into an embodied policy that succeeds under physical task-completion criteria.
    The benchmark contains 32 simulation tasks across four platforms and two development regimes: improving a functional human-authored codebase and completing a policy from skeletal scaffolding.
    We compare three conditions: the platform-provided human expert reference, a non-agentic one-shot LLM condition, and the full \textsc{RoboCoach} loop.
    All results are evaluated only by binary task success.
    Details of task construction, model coverage, aggregation rules, failure handling, and real-robot transfer protocols are provided in Methods; full per-task and per-model values are reported in Extended Data Tables~\ref{tab:benchmark_performance_improved}--\ref{tab:modelwise_performance} and the complete benchmark table (Supplementary Table~1).
    
    \begin{figure}[htbp]
        \centering
        \includegraphics[width=0.95\textwidth]{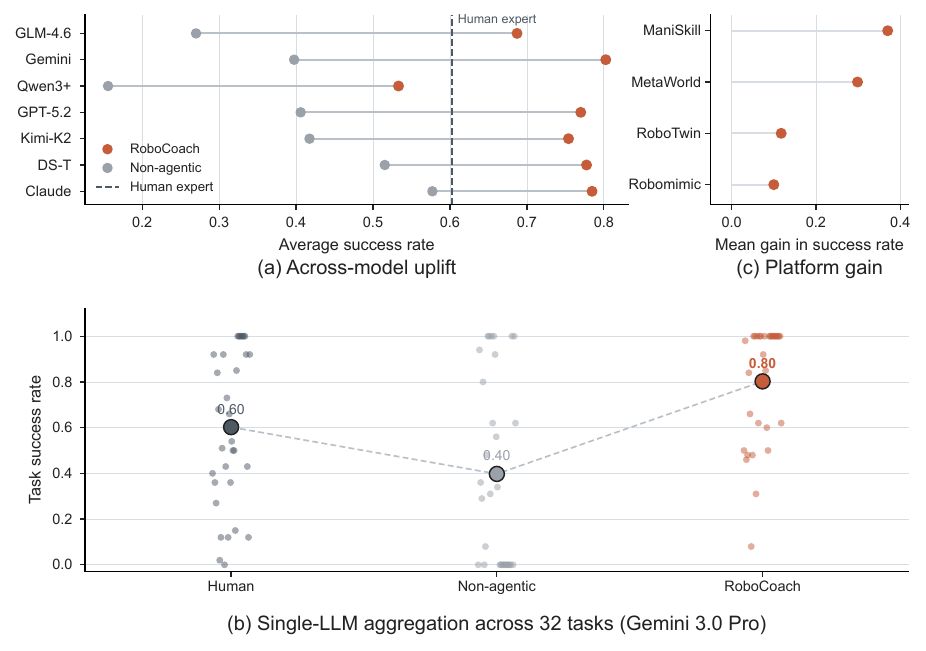}
        \captionsetup{justification=RaggedRight, singlelinecheck=false}
        \caption{\textbf{Overall effectiveness across base LLMs, evaluation settings, and embodied platforms.}
        \textbf{a,} Model-wise uplift from non-agentic prompting to \textsc{RoboCoach}; the dashed line marks the 32-task human expert reference.
        \textbf{b,} Single-LLM benchmark summary using Gemini~3.0~Pro, with task points and aggregate means for human expert, non-agentic, and \textsc{RoboCoach} conditions.
        \textbf{c,} Platform-wise gain of \textsc{RoboCoach} over the human expert reference under the same single-LLM aggregation.}
        \label{fig:main_results}
    \end{figure}
    \subsection{Overall effectiveness of RoboCoach}
    \label{subsec:overall}
    We first ask whether frontier language models can achieve embodied competence through one-shot generation alone.
    Across \textsc{RoboCoach-Bench}, they do not.
    Non-agentic prompting often produces runnable code and, in some cases, trainable pipelines, but these outputs remain unreliable under the physical task-completion criterion.
    Activating the full \textsc{RoboCoach} loop substantially reduces this embodied competence gap.
    This distinction matters because the non-agentic condition already gives the model the task specification and codebase; the remaining gap is therefore not simply a lack of task description, but a failure to connect generated code to physical consequences.
    In the improving setting, among model--task pairs where both conditions completed, 60.5\% improved under \textsc{RoboCoach}, 39.5\% were unchanged, and none declined.
    Failed one-shot runs were also frequently recovered by the closed-loop system, indicating that part of the gain comes from turning brittle generated artefacts into executable and physically evaluated policies.
    
    Across the seven evaluated base models, all non-agentic model-wise averages remain below the human expert reference, whereas \textsc{RoboCoach} raises the benchmark-wide mean to 0.730, above the human expert mean of 0.602.
    Six of the seven base models exceed the human reference after closed-loop optimisation, indicating that the effect is not tied to a single unusually strong model.
    Under the single-LLM aggregation used for the main benchmark summary, performance rises from 0.40 for non-agentic prompting to 0.80 with \textsc{RoboCoach}, compared with 0.60 for the human expert reference.
    
    This gap closure is consistent across initial-code regimes and benchmark families.
    \textsc{RoboCoach} improves performance in both the improving and from-scratch settings, and the aggregate advantage over the human reference is positive across all four simulation platforms.
    The from-scratch setting should be interpreted with this structure in mind: several tasks already have platform references near ceiling, so the more informative contrast is the recovery from weak or failed one-shot development to physically successful policies under the same evaluation criterion.
    Detailed per-model, per-setting, and task-oracle values are reported in Extended Data Tables~\ref{tab:benchmark_performance_improved}--\ref{tab:modelwise_performance}.
    
    \subsection{Task-wide distribution of gains}
    \label{subsec:task_distribution}
\begin{figure}[htbp]
        \centering
        \includegraphics[width=\textwidth]{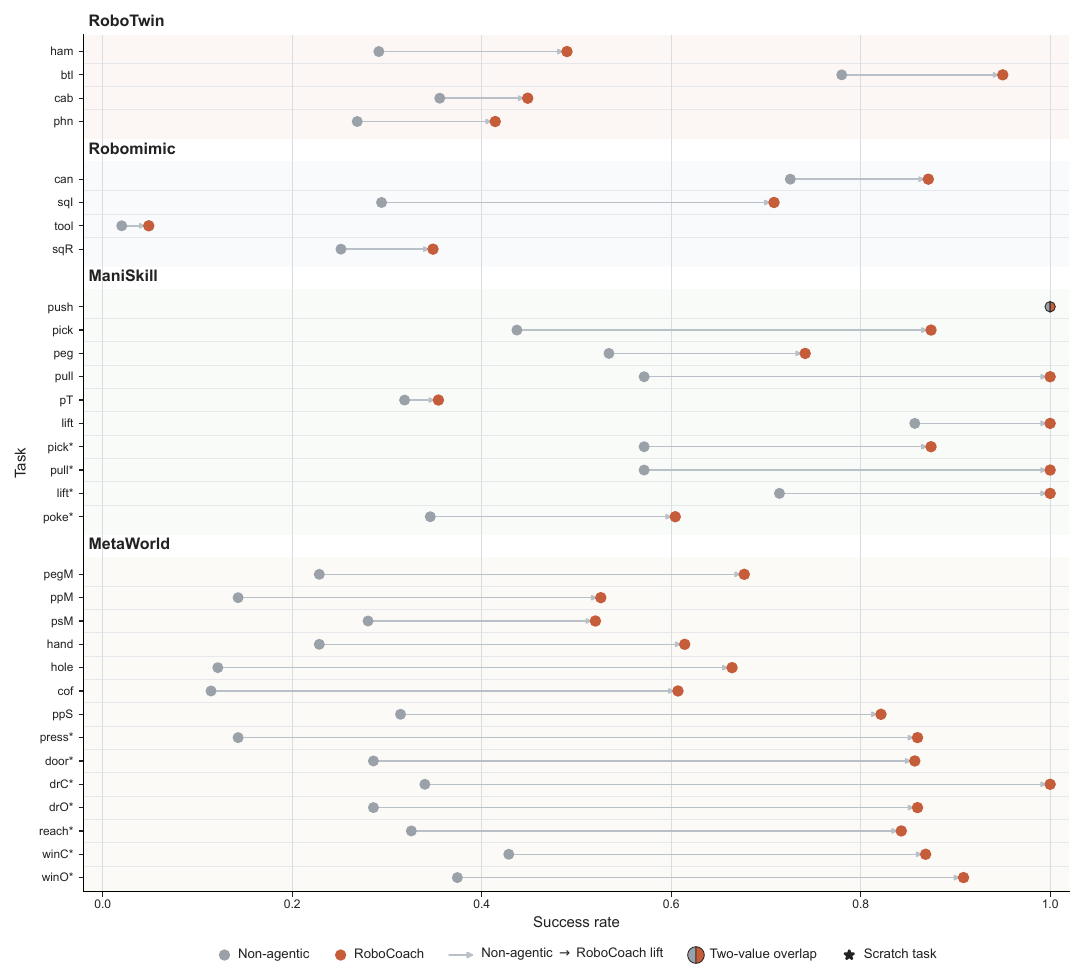}
        \captionsetup{justification=RaggedRight, singlelinecheck=false}
        \caption{\textbf{Task-wide distribution of gains.}
       Per-task success rates for non-agentic prompting and \textsc{RoboCoach}, grouped by platform and averaged over the seven base models. Arrows show the per-task lift; scratch tasks are marked with a trailing \texttt{*}. Task abbreviations are defined in Extended Data Table~\ref{tab:task_abbreviations_metadata}.}
        \label{fig:task_distribution}
    \end{figure}

    We next ask whether the aggregate improvement is broadly distributed across tasks or driven by a small set of favourable cases.
    Fig.~\ref{fig:task_distribution} reports the seven-model mean success rate for each task under non-agentic prompting and \textsc{RoboCoach}.
    The distribution shifts upward across the benchmark: improvements appear across simulation platforms, learning paradigms, and policy architectures, including reinforcement learning, imitation learning, MLP policies, recurrent policies, diffusion policies, action chunking transformers, and vision-language-action policies.
    
    The size of the gain is structured by the headroom left by one-shot generation.
    Tasks where non-agentic prompting already approaches the ceiling show smaller improvements, whereas tasks where one-shot generation fails to produce physically successful behaviour show larger gains.
    This pattern is visible both in reinforcement-learning tasks, where code repair, reward shaping, and optimisation choices can rapidly change success rates, and in imitation-learning tasks, where policy architecture, data processing, and rollout behaviour constrain the attainable gain.
    At the same time, the distribution does not collapse to perfect success.
    A small residual frontier of hard tasks remains even under closed-loop optimisation, especially where contact-rich or visually mediated behaviours leave little room for simple code-level repair, motivating the mechanistic analysis below.
    Full task names, abbreviations, and per-task values are reported in Extended Data Table~\ref{tab:task_abbreviations_metadata} and the complete benchmark table (Supplementary Table~1).
    
    \subsection{Mechanistic sources of improvement and remaining difficulty frontier}
    \label{subsec:ablation}
    We then examine which parts of the closed-loop framework account for the improvement.
    The key issue is not only which module contributes most in aggregate, but what type of embodied failure becomes diagnosable when that module is present.
    The ablation study shows that removing any feedback channel or the branching search mechanism reduces success relative to the full system.
    Textual execution feedback and branching search produce the largest drops, consistent with their roles in resolving implementation failures and escaping locally plausible but physically ineffective trajectories.
    Video rollout observations and quantitative training signals contribute complementary gains by exposing behavioural and optimisation-level failures that are not reliably visible from code execution alone.
    On the four-task RoboTwin ablation set, the full system reaches a mean success rate of 0.61.
    Removing textual execution signals reduces this mean to 0.50, and removing branching search reduces it to 0.52, while removing video observations or quantitative training signals gives smaller but still measurable drops.
    These numbers support a layered interpretation: the system first needs to make the code run, then choose which physical hypotheses to continue, and finally refine the policy using training dynamics and behaviour-level evidence.
    
    \begin{figure}[htbp]
        \centering
        \includegraphics[width=\textwidth]{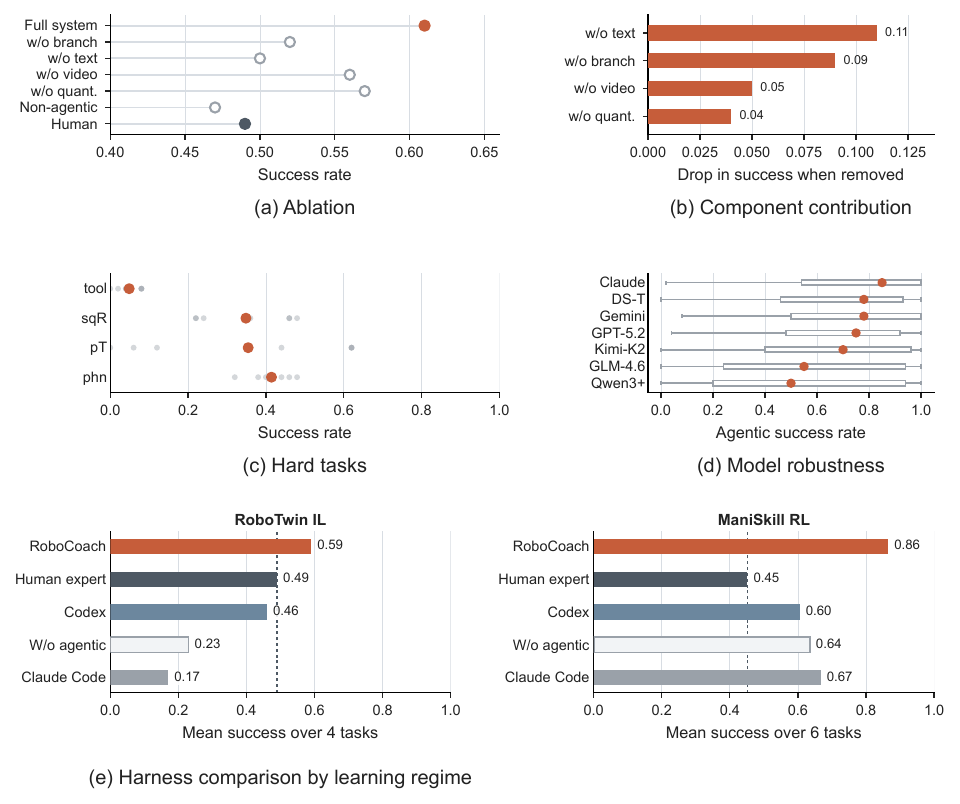}
        \captionsetup{justification=RaggedRight, singlelinecheck=false}
        \caption{\textbf{Mechanistic contribution of feedback components and residual heterogeneity.}
        \textbf{a,} Mean success rate of the full system, component-removal variants, and reference baselines.
        \textbf{b,} Performance drop from removing each component.
        \textbf{c,} Residual task difficulty across base models.
        \textbf{d,} Per-model distribution over improving tasks.
        \textbf{e,} Matched-model comparison with generic coding-agent harnesses under GPT-5.2.}
        \label{fig:mcts_search}
    \end{figure}
    
    The same analysis also reveals a residual hard-task frontier.
    Some tasks remain difficult across all seven base models even with \textsc{RoboCoach}, showing that benchmark-wide improvement and persistent task-level heterogeneity coexist.
    
    Finally, we compare \textsc{RoboCoach} with generic coding-agent workflows under a fixed base model.
    Under this matched-model setting, \textsc{RoboCoach} remains the strongest tested harness in both the imitation-learning and reinforcement-learning subsets, suggesting that the advantage comes from the embodied feedback-and-search structure rather than from base-model strength alone.
    With GPT-5.2 fixed as the base model, \textsc{RoboCoach} reaches 0.59 on the RoboTwin imitation-learning subset, above the human reference, Codex, Claude Code, and non-agentic prompting.
    On the ManiSkill reinforcement-learning subset, it reaches 0.86, again exceeding the matched generic coding-agent workflows.
    Exact ablation values and harness-comparison means are reported in Fig.~\ref{fig:mcts_search} and Extended Data Table~\ref{tab:ablation_info_sources}.
    \FloatBarrier
    
    \begin{figure}[htbp]
        \centering
        \includegraphics[width=0.98\linewidth]{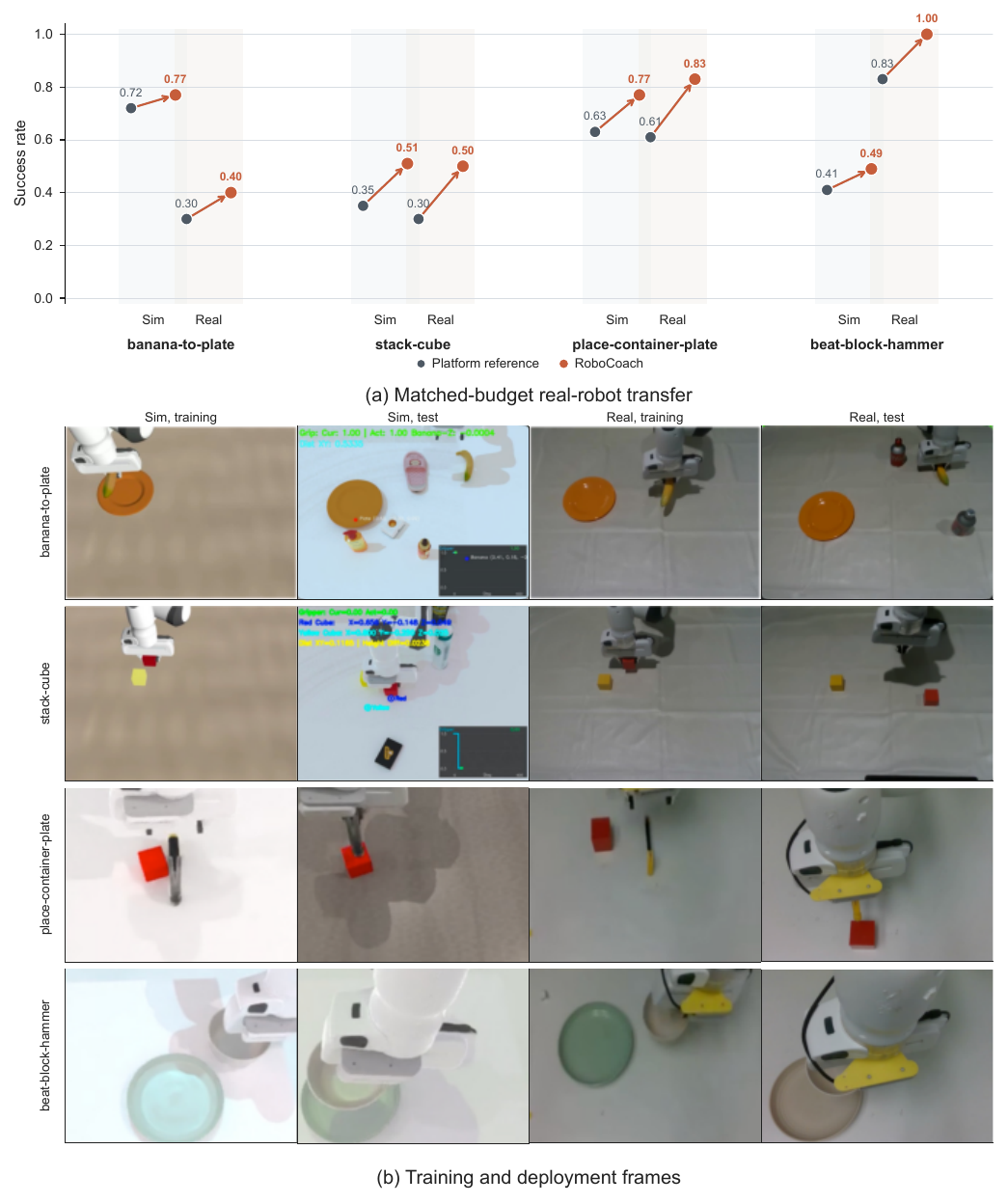}
        \captionsetup{justification=RaggedRight, singlelinecheck=false}
        \caption{\textbf{Real-robot transfer across two hardware settings.}
        \textbf{a,} Simulation and real-robot success rates for four transfer tasks under matched budgets.
        \textbf{b,} Representative training and deployment frames.}
        \label{fig:sim2sim2real}
    \end{figure}

    \subsection{Real-robot transfer across hardware settings}
    \label{subsec:realrobot}
    
    We finally test whether the simulation gains reflect physically meaningful competence rather than simulator-specific optimisation.
    Across four real-robot transfer tasks in two hardware laboratories, policies optimised by \textsc{RoboCoach} preserve their advantage over matched platform references after deployment on physical robots.
    Averaged over the four transfer tasks, \textsc{RoboCoach} improves the simulation success rate from 0.53 to 0.64 and the real-robot success rate from 0.51 to 0.68.
    
    The improvement appears under both transfer regimes: mixed simulation--real finetuning on the first hardware platform and real-only finetuning on the second.
    This suggests that the closed-loop development process captures task-relevant structure that survives transfer to physical hardware, rather than merely exploiting simulator-specific reward or evaluation artefacts.
    The transfer result is deliberately modest in scope, but it is important for attribution: policies developed under the same task definitions and budget constraints retain their advantage after deployment, including when the agent receives physical rollout feedback from real hardware during finetuning.
    Task-level success rates, rollout counts, and hardware-specific protocols are reported in Methods and Extended Data Table~\ref{tab:real_robot_transfer_results}.
    \FloatBarrier
    
    \section{Discussion}
    This study asks whether the growing competence of digital agents can be converted into physical competence: can an agent that acts only through code, tools and mediated feedback develop policies that succeed under embodied task-completion criteria? Within the bounded settings tested here, our results give a positive but qualified answer. Across RoboCoach-Bench and complementary real-robot transfer experiments, RoboCoach shows that digital agents can improve physical task outcomes when their code-level interventions are embedded in a loop that measures physical consequences, interprets failure and revises the policy-development process. The important distinction is therefore not between language and embodiment alone, but between generating a digital artefact and closing a development loop whose objective is physical task completion. One-shot code generation often produces runnable or trainable systems, yet remains unreliable as a route to embodied competence.
    
    The results also identify the mechanism behind this transition. The improvement is not explained by model capability alone: the same base models perform substantially differently when deprived of execution feedback, behavioural observations and feedback-guided search. RoboCoach instead gains from organising three kinds of evidence around each attempted policy: textual signals that expose implementation failures, quantitative signals that expose optimisation failures, and rollout-video diagnostics that expose behavioural failures only visible during physical execution. Tree-structured memory and branching search then turn these attempts into an auditable sequence of physical hypotheses, allowing the agent to extend promising branches, roll back from locally plausible failures or restart from stronger states. In this sense, the agent does not need direct sensory embodiment to benefit from physical grounding; it needs a system architecture that makes physical consequences legible and actionable.
    
    The logged modification trajectory in Supplementary Information Section~6 also clarifies why this loop should not be reduced to AutoML-style hyperparameter search.
    In the representative diffusion-policy run, \textsc{RoboCoach} changes the training system at several levels: it enables gradient checkpointing and mixed-precision training to make larger models feasible, adds gradient clipping and EMA settings to stabilise optimisation, increases model capacity and temporal horizon, changes learning-rate schedules, adds image and state augmentation, smooths action sequences, fixes validation-mode logic, corrects validation postprocessing, and repairs checkpointing and runtime failures.
    These are not independent scalar knobs around a fixed learner.
    They are coordinated interventions across model architecture, data handling, optimisation, validation, and infrastructure, each evaluated by its downstream effect on physical task completion.
    This breadth is central to the claim: \textsc{RoboCoach} automates a bounded version of the robotics research loop, not only the search over a predefined hyperparameter space.
    
    Placed against prior work, this points to a different unit of automation. Earlier language-agent systems for embodied AI have automated important components, including reward design, domain randomisation, control synthesis, task generation and simulator construction~\cite{Eureka,text2reward,DrEureka,codeaspolicy,RoboGen,gensim}. These are valuable Level-1 capabilities in the roadmap of Fig.~\ref{fig:cover}b. RoboCoach targets Level 2: the autonomous development or improvement of a complete policy under a fixed task definition and evaluation protocol. This distinction matters because the bottleneck in routine embodied-AI research is rarely a single missing component. It is the repeated cycle in which a researcher writes code, launches training, inspects failures, changes the policy or training setup, and decides whether to continue or abandon an attempt. The present results suggest that this loop can itself become an object of automation.
    
    RoboCoach-Bench is designed to make that claim measurable. Rather than evaluating only the final policy or the coding ability of an LLM, it evaluates whether an agentic development process can convert a natural-language task specification into a policy that completes the task. The use of a single physical task-completion metric, paired expert-authored references and matched non-agentic conditions helps separate three sources of progress: the prior knowledge of the base model, the strength of the platform reference and the contribution of the closed-loop framework. At the same time, the expert reference should not be interpreted as an upper bound on human capability; it is a controlled comparison point for measuring whether an autonomous development loop can improve upon strong platform-provided configurations. This evaluation principle is especially important for embodied AI, where progress in code quality, loss curves or reward magnitude can be mistaken for physical competence. The benchmark should therefore be read not only as an evaluation of RoboCoach, but as a step towards measuring autonomous embodied engineering as a capability in its own right.
    
    The conclusion is bounded in several ways. First, the main evidence remains simulation-based, and the real-robot experiments cover four tasks across two hardware settings rather than the diversity of morphologies, sensors and deployment conditions encountered in open-world robotics. Second, the role of embodied feedback is task-dependent. In contact-rich manipulation, imitation learning and cross-domain transfer, rollout observations can reveal failures that scalar metrics miss; in simpler reinforcement-learning tasks, part of the gain may instead come from conventional code repair, reward shaping or hyperparameter search. Third, binary task success provides a clear grounding signal but does not fully measure sample efficiency, robustness under distribution shift, near-miss behaviour or safety-critical failures. Fourth, branching search improves reliability at non-trivial computational cost. Finally, RoboCoach assumes well-formed task specifications, fixed evaluation protocols and bounded edit interfaces. These constraints are useful for scientific attribution, but they are far from the uncertainty, non-stationarity and safety requirements of deployed robot systems.
    
    These limitations define the next stage rather than weakening the central claim. Moving beyond autonomous single-policy development will require larger-scale hardware-in-the-loop evaluation, budget-aware search, cross-task memory transfer, richer multi-objective metrics and safety checks that are integrated into the optimisation loop rather than added after deployment. More broadly, the results suggest that progress in embodied AI may depend increasingly on how foundation models are organised within systems that structure feedback, memory, search, data acquisition and physical validation. RoboCoach does not demonstrate autonomous development of embodied foundation models or open-world robot agents. It establishes a bounded Level-2 setting in which digital agents can be evaluated as developers of physical competence, and it makes the requirements for advancing toward Level 3 and Level 4 more concrete.

\section{Methods}\label{sec:methods}

The framework overview is shown in Fig.~\ref{fig:framework_and_benchmark}. The two obstacles identified in the Introduction set the structure of this section. To address the \emph{methodological} obstacle, that embodied policy development requires diagnosing physical failure from behavioural observation, we introduce \textsc{RoboCoach}, an autonomous framework that lets a purely digital LLM-based agent improve embodied policies through closed-loop interaction with embodied environments. Its three components are developed in turn: a task-tuple formalisation that bounds what the agent is permitted to know and do (Sec.~\ref{subsec:formalisation}), a physically grounded feedback mechanism that channels execution outcomes back into policy revision across three complementary signal classes (Sec.~\ref{subsec:feedback}), and a tree-structured memory that supports feedback-guided branching search over iterations (Sec.~\ref{subsec:memory}). To address the \emph{evaluation} obstacle, that physical competence cannot be assessed at scale on real robots alone, we introduce \textsc{RoboCoach-Bench} and its evaluation protocol, which pair scalable closed-loop simulation with real-robot transfer (Secs.~\ref{subsec:benchmarkdesign} and~\ref{subsec:evalprotocol}).

\subsection{Task-tuple formalisation enables
            bounded digital-to-physical optimisation}\label{subsec:formalisation}

To make the central question concrete and testable, we formalise each
engineering task as a tripartite tuple:
\begin{equation}
  \mathcal{T}
  = \bigl(\mathcal{D}_\mathrm{prd},\;
          \mathcal{P}_\mathrm{sys},\;
          \mathcal{C}_\mathrm{env}\bigr),
  \label{eq:task_tuple}
\end{equation}
where the three components together define the boundary of what the agent is permitted to know and do. The first component, $\mathcal{D}_\mathrm{prd}$ (Semantic Specification), is a structured natural-language document encoding the optimisation objective, operational constraints (resource budgets, immutable evaluation metrics, restricted file access), and physics-informed domain priors; critically, it provides \emph{no} demonstrations, reward gradients, or labelled trajectories; every learning signal must be earned through embodied execution in the simulator or robot setup. The second component, $\mathcal{P}_\mathrm{sys}$ (Operational Interface), equips the agent with a \texttt{Terminal}, a \texttt{FileEditor}, and a \texttt{TaskTracker}; all interventions flow through code manipulation with no direct sensor or actuator access. The third component, $\mathcal{C}_\mathrm{env}$ (Development Substrate), is the execution environment: either a functional but sub-optimal human-authored codebase (improving setting) or a skeletal template with core logic omitted (from-scratch setting); the sole dependent variable is \emph{task success rate}, with no proxy metric permitted.
This task tuple formalises the embodied competence gap introduced in the Introduction: the agent acts only through digital interfaces, represented by $\mathcal{P}_\mathrm{sys}$, while success is defined only by physical outcomes in $\mathcal{C}_\mathrm{env}$, and the gap is precisely the span between these two levels that the agent cannot directly observe.
In our formulation, embodied competence is therefore not assumed a priori, but measured by whether physically grounded feedback can reliably transform digital actions into improved task success.

\subsection{Physically grounded feedback anchors
            policy optimisation in physical reality}\label{subsec:feedback}

One-shot generation is particularly brittle for embodied tasks because executable code is not equivalent to physical competence.
A policy may compile, launch, and even exhibit apparently stable optimisation dynamics while still failing at the level of contact, control, or task completion in the environment.
\textsc{RoboCoach} addresses this physical grounding gap through three complementary feedback channels, each resolving a different class of failure that is invisible at the others' level of abstraction and translating grounded diagnoses into policy updates.
Prior works~\cite{Eureka,text2reward,DrEureka} addressed only well-scoped sub-tasks with text-only or
reward-only feedback.

\paragraph{Textual execution signals.}
Textual execution signals provide the first layer of grounding by exposing implementation failures, such as syntax errors, runtime exceptions, misconfigured dependencies, and invalid experiment logic. They are necessary because an agent cannot improve a policy that does not run, but they are not sufficient for embodied competence: many implementations execute correctly while still producing physically ineffective behaviour. To keep this layer cheap, a lightweight \texttt{debug\_test} tool submits each modified codebase to a short 10-episode validation run before committing to a full cluster job, catching implementation failures well before they consume training resources.

\paragraph{Quantitative training signals.}
Quantitative training signals provide a second layer of grounding by exposing optimisation failures, such as unstable learning dynamics, reward collapse, stalled improvement, or ineffective hyperparameter settings. These signals are themselves not new; what matters is that the agent treats them as a first-class observation stream rather than waiting for the system to surface them. \textsc{RoboCoach} equips the coding agent with experiment-tracking tools (for example, Weights \& Biases) so that it autonomously instruments each run, queries its own reward curves, loss trajectories, and running success rates on demand, and decides when a run is learning, diverging, or plateauing. This lets the agent distinguish non-learning from learning and revise the training process accordingly, yet scalar progress alone is still insufficient, because learning curves can improve while the resulting behaviour remains physically misaligned with the task objective, motivating the third channel below.

\paragraph{Video rollout observations.}
The third channel is perception: the agent observes rollout videos of the trained policy acting in the environment. A vision-language model converts each rollout video into a compact structured summary of the observed behaviour, which is the diagnostic form in which this channel surfaces failure. Using a vision-language model as an interpreter of embodied rollouts follows a growing line of work that treats such models as zero-shot judges or reward sources for robot behaviour~\cite{baumli2024vlm,ma2024vlmrm}; here it is repurposed to produce structured behavioural diagnostics rather than scalar rewards. Through this channel, the agent can diagnose behavioural failures that are often invisible to logs and scalar metrics, such as unstable contact, incorrect grasp posture, inefficient motion, or trajectory deviations near the goal state. Such failures are especially consequential for expressive policy classes such as diffusion- and flow-based manipulation policies~\cite{chi2025DP,zhang2025flowpolicy}, where smooth training curves can coexist with physically brittle behaviour.
In this way, the agent is grounded not only in whether code runs or training progresses, but in how the policy actually behaves in the physical task loop. Concretely, after each full training run and subsequent validation rollout, rollout videos are analysed by a vision-language model that produces a structured natural-language summary, describing for example that ``the robot arm exhibits jerky, discontinuous motion during the approach phase, causing the object to be knocked aside rather than grasped''. This behavioural summary is then fed back into the next iteration as a physics-informed diagnostic.

Taken together, these three signal classes ground policy-failure understanding in physical reality rather than in code inference alone, which is the key ingredient distinguishing \textsc{RoboCoach} from prior code-generation approaches to embodied AI.
They form a hierarchy of physical grounding in which each channel diagnoses a distinct failure class: textual execution signals diagnose implementation failures, quantitative training signals diagnose optimisation failures, and video rollout observations diagnose behavioural failures that emerge only in physical execution.

\subsection{Tree-structured memory enables
            feedback-guided search}\label{subsec:memory}

Embodied policy development is not only a coding problem but a search over physical hypotheses, in which each attempt leaves behind outcomes that later attempts must be able to inherit from, diverge from, or discard; a naive transcript-based agent cannot support this, because it must either forget earlier rounds or refuse to explore branches diverging from its current attempt.
\textsc{RoboCoach} replaces that flat history with a tree-structured memory of physically evaluated attempts and a pair of agents that operate on it, providing the memory and search components of the framework that complement the physically grounded feedback of Sec.~\ref{subsec:feedback}. This design echoes classical tree-search approaches to sequential decision-making~\cite{kocsis2006uct,browne2012mcts} and their recent language-model analogues for structured reasoning and tool use~\cite{zhou2024lats}, but replaces symbolic or textual rewards with measured physical task success at each node.

\paragraph{Round-level isolated workspaces.}
Each iteration round operates in a fully isolated codebase directory.
The workspace isolation module copies only source files into each
round's directory while mounting large assets (datasets, model
weights, simulator binaries) via read-only links, avoiding redundant
data duplication across rounds.
Every round therefore has a reproducible, independently auditable
snapshot of the code state at that point in the search.

\paragraph{Tree-structured memory.}
Each node in the tree represents a complete embodied experiment rather than a code snapshot alone: it stores the implementation state together with execution outcomes, training dynamics, video rollout observations, and measured task success. Directed edges encode inheritance, recording how later attempts are initialised from or diverge from earlier ones. This tree is therefore the agent's persistent memory of its own exploration history, preserving not only what was changed, but what physical consequences those changes produced.

\paragraph{Dual-agent loop with feedback-guided branching.}
The two agents alternate in a closed \emph{Draft, Debug, and Improve} cycle with a strict division of labour.
The coding agent performs local edits within a selected branch, whereas the feedback agent decides where the next edit should be rooted.
After each job finishes, the feedback agent reads the structured round record, including job status, metrics, log tail, and rollout-video summary, and produces both a compact recommendation for the next edit and a parent choice for the next search step.
The parent choice can extend the current branch, roll back to the best measured node, or restart from the baseline.
This makes branching depend on physical evidence rather than recency: a branch with improving training curves can be abandoned if rollout observations reveal unstable contact or incorrect task execution, and a stalled branch can be replaced by the current global best.
Extended Data Fig.~\ref{fig:ed_search_graph} illustrates this search graph for a representative run.
It is not used as an additional performance result; rather, it documents how \textsc{RoboCoach} turns sequential trial-and-error into auditable search, with nodes denoting physically evaluated attempts and edges denoting inheritance between attempts.

\subsection{Benchmark design: \textsc{RoboCoach-Bench}}\label{subsec:benchmarkdesign}

To answer whether \textsc{RoboCoach} can close the embodied competence gap across diverse embodied-engineering settings, we require a benchmark that (i) is broad enough to test
generalisation across learning paradigms and policy architectures,
(ii) is grounded in physical task completion rather than code-quality
proxies, and (iii) provides a paired human expert baseline against which
gains can be attributed.
\textsc{RoboCoach-Bench} is designed to satisfy all three criteria; an overview of its composition is given in the task catalogue of Sec.~\ref{subsec:benchmarkdesign} and the full performance tables (Extended Data Tables~\ref{tab:benchmark_performance_improved} and~\ref{tab:benchmark_performance_scratch}).

\paragraph{Platform and task coverage.}
The benchmark comprises 32 expert-curated tasks drawn from four
high-fidelity simulation platforms: ManiSkill3~\cite{maniskill3},
RoboTwin2~\cite{robotwin2}, Robomimic~\cite{robomimic}, and
MetaWorld~\cite{metaworld}.
These platforms were selected for their extensive community adoption,
robust physics engines, and complementary task diversity, ranging
from bimanual manipulation and contact-rich assembly to multi-task
RL curricula.

\paragraph{Learning paradigm and architecture diversity.}
Tasks span both reinforcement learning and imitation learning,
and cover policy architectures from multilayer perceptrons and
recurrent networks~\cite{robomimic} to Diffusion Policies~\cite{chi2025DP}, Action
Chunking Transformers~\cite{zhao2023ACT}, and vision-language-action
models~\cite{openvla}.
This diversity is deliberate: the benchmark is designed to probe
whether the gains from \textsc{RoboCoach} are architecture-agnostic
or confined to specific policy classes.

\paragraph{Evaluation settings.}
In the \emph{improving setting} (21 tasks), the agent starts from a
functional but sub-optimal platform-provided human-authored codebase.
In the \emph{from-scratch setting} (11 tasks), the agent receives
only simulator bindings and high-level templates with core logic
omitted, requiring it to complete the training pipeline from minimal
scaffolding.
Each task is paired with a natural-language specification
($\mathcal{D}_\mathrm{prd}$) and a platform-provided human-authored baseline solution.

\subsection{Evaluation protocol}\label{subsec:evalprotocol}

\paragraph{Platform reference baseline.}
For each task, the human expert baseline records the task success rate
achieved by the platform-provided human-authored codebase submitted as part of the
benchmark, evaluated over 100 episodes.
These implementations constitute strong expert-authored baselines that reflect standard practice within each platform and have typically been refined through repeated community use and benchmarking.
Our comparison therefore measures whether the agent can improve upon a strong expert reference point rather than a naive first attempt.
At the same time, this baseline should not be interpreted as an upper bound on human capability; it is a rigorous reference configuration, not a claim about the best possible human performance under unrestricted effort.

\paragraph{Attribution conditions.}
Three conditions isolate the source of performance gains.
The \emph{human expert baseline} provides the pre-agent reference by using the platform-provided human-authored codebase as the reference solution.
The \emph{non-agentic} condition prompts the LLM once with the
task specification and the full codebase, without any execution
feedback, to measure the contribution of raw model knowledge.
The \emph{agentic} condition activates the full
\textsc{RoboCoach} loop, including physically grounded feedback, tree-structured
memory, and dual-agent branching, to measure the additional
contribution of the closed-loop framework.
The video-feedback analysis tool in the agentic condition is implemented with Qwen3-VL-235B-A22B-Instruct~\cite{Qwen3-VL}.
Comparing non-agentic to agentic performance directly answers the
central design claim: that the gains of \textsc{RoboCoach} arise
from the framework, not from the model's pre-trained coding knowledge.

\paragraph{Model coverage.}
Seven frontier language models are evaluated across both conditions:
Claude Opus~4.5, Gemini~3.0 Pro, GPT-5.2,
DeepSeek V3.2-Thinking, Kimi-K2-Thinking,
Qwen3-Coder-Plus, and GLM-4.6.
This set spans both instruction-tuned and reasoning-specialised
models, enabling analysis of how model capability interacts with
the agentic framework.

\paragraph{Aggregation and reporting.}
We use three aggregation protocols for distinct purposes.
First, \emph{model-wise averages} compute each model's mean success rate across all 32 benchmark tasks, enabling cross-model comparison.
Second, \emph{single-LLM aggregation} fixes one base model for every task, chosen as the model with the highest benchmark-wide \textsc{RoboCoach} mean, and is used for the main aggregate comparison in Fig.~\ref{fig:main_results}b--c.
Third, \emph{task-oracle averages} select the best-performing model separately for each task and are reported only in Extended Data Tables~\ref{tab:benchmark_performance_improved} and~\ref{tab:benchmark_performance_scratch}.
Entries marked \textit{bug} denote failed runs and count as zero success in aggregate calculations.

\paragraph{Sole success criterion.}
For all tasks and all conditions, the only recorded metric is
\emph{task success rate}.
For simulation benchmark tasks, this is the fraction of 100 evaluation episodes
in which the trained policy achieves the binary physical completion
criterion defined by the simulator.
For real-robot transfer tasks, this is the fraction of physical evaluation rollouts
that satisfy the corresponding real-world completion criterion, with rollout counts specified in the transfer protocol below.
No proxy metric (code quality, reward magnitude, or human judgment)
substitutes for this ground truth.
This design ensures that any improvement recorded by the benchmark
reflects embodied engineering competence under the corresponding simulator or real-robot completion criterion.
It also reduces the risk that apparent progress in digital space is mistaken for embodied competence when the resulting policy does not actually solve the physical task.

\paragraph{Real-robot transfer protocol.}
To test whether embodied-development gains extend to physical hardware while reducing dependence on a single robot setup, we evaluate the four real-robot tasks in two hardware laboratories under two matched data regimes. In the first setting, used for banana-to-plate and stack-cube, we run the Sim2Sim2Real VLA-finetuning protocol: simulation training data are generated in Genesis~\cite{genesis}, simulation evaluation uses RoboTwin~\cite{robotwin2} under randomised conditions, and real-robot evaluation uses the physical arm with distractor objects added to the scene.
Finetuning data in this setting are deliberately imbalanced toward simulation (for example, 300 Genesis trajectories against 30 real-robot trajectories for banana-to-plate), so that the real-robot evaluation measures the physical transferability of simulation-optimised behaviour rather than the effect of large-scale real-world data collection. Both the platform reference and \textsc{RoboCoach} start from the same $\pi_{0.5}$~\cite{intelligence2025pi05} policy, train for nine hours on eight RTX 4090 GPUs under identical pre-deployment restrictions, and are evaluated using the latest checkpoint at the training deadline. Real-robot success is measured over 10 physical evaluation rollouts.
In the second setting, used for place-container-plate and beat-block-hammer, we run an independent real-robot experiment in a second laboratory on a Franka Panda arm. These two tasks use real-only finetuning data, with 90 real-robot trajectories and no simulation trajectories in the finetuning set, and real-robot success is measured over 18 physical evaluation rollouts.
During finetuning in both settings, the agent receives physical rollout observations, including wrist-camera video, as additional feedback; at each iteration it proposes configuration updates, launches the next training run, and retains the best-performing checkpoint. The resulting real-robot policy is evaluated under the same binary completion criterion used elsewhere in the benchmark and is compared against a human-engineered real-robot configuration prepared under matched task definitions and compute budgets.

\subsection*{Data availability}
Benchmark task specifications, evaluation protocols, and aggregated performance data are provided with the submission for editorial and peer-review assessment.
The public release will be made available upon publication at a persistent repository.
Source data underlying the main figures and Extended Data tables will be provided with the final accepted version.

\subsection*{Code availability}
Custom code central to the reported results is available to editors and reviewers for assessment during peer review through a private repository.
The full codebase will be released upon publication at a persistent public repository.

\subsection*{Acknowledgements}
The authors thank the developers of ManiSkill, RoboTwin, Robomimic,
MetaWorld and OpenHands for making their platforms publicly available.

\section*{Declarations}
\begin{itemize}
\item \textbf{Funding} This work was supported by Zhongguancun Academy (Grant No. 02012405).
\item \textbf{Competing interests} The authors declare no competing interests.
\item \textbf{Ethics approval} Not applicable.
\item \textbf{Consent for publication} All authors consent to publication.
\item \textbf{Author contributions}
Zixing Lei and Siheng Chen led the project.
Zixing Lei, Genjia Liu, and Yuanshuo Zhang developed the system framework and conducted the main simulation experiments.
Qipeng Liu, Sixiang Chen, Jixian Wu, Shanghang Zhang, and Wenzhao Lian supported the real-robot experiments across two laboratories.
Chuan Wen, Shanghang Zhang, Wenzhao Lian, Yunhong Wang, and Weixin Li provided conceptual guidance and contributed to manuscript writing and revision.
\end{itemize}

\clearpage
\newpage
\bibliography{main}
\newpage
\section{Extended Data}

\setcounter{figure}{0}
\setcounter{table}{0}
\renewcommand{\thefigure}{\arabic{figure}}
\renewcommand{\thetable}{\arabic{table}}
\captionsetup[figure]{labelformat=empty}
\captionsetup[table]{labelformat=empty}

\begin{table}[htbp]
\caption{\textbf{Extended Data Table~1 $|$ Benchmark tasks and agent
performance in the improving setting.}
``w/o agentic'': single-shot generation without execution feedback.
``agentic'': iteratively optimised generation via RoboCoach.
\textbf{Aggregation rule:} ``Best LLM'' is selected per task by highest
\textit{agentic} success; averages are task-oracle aggregates, not
model-wise means.
\textbf{Bug handling:} entries marked \textit{bug} are failed runs and
count as 0.00 success when computing averages.
Values: task success rate (fraction of 100 episodes).
Superscript $^{a}$ denotes the MT-10 multi-task setting.
Superscript $^{b}$ denotes individual single-task environments.}
\label{tab:benchmark_performance_improved}
\centering

\fontfamily{ptm}\selectfont
\setlength{\tabcolsep}{2pt}
\fontsize{7.5pt}{9pt}\selectfont

\begin{tabular}{lllllcll}
\toprule
\textbf{Task} & \textbf{Best LLM} & \textbf{Env.} & \textbf{Mode} &
\textbf{Arch.} & \textbf{Human} &
\textbf{w/o agentic} & \textbf{agentic} \\
\midrule
beat-block-hammer          & Claude Opus 4.5        & RoboTwin  & IL & ACT       & 0.40
  & 0.40 \textcolor{BrickRed}{(+0.00)}
  & 0.54 \textcolor{BrickRed}{(+0.14)} \\
adjust-bottle              & Gemini 3.0 Pro         & RoboTwin  & IL & ACT       & 0.92
  & 0.94 \textcolor{BrickRed}{(+0.02)}
  & 0.98 \textcolor{BrickRed}{(+0.06)} \\
put-object-cabinet         & Kimi-K2-Thinking       & RoboTwin  & IL & Diffusion & 0.36
  & 0.34 \textcolor{OliveGreen}{($-$0.02)}
  & 0.54 \textcolor{BrickRed}{(+0.18)} \\
place-phone-stand          & Gemini 3.0 Pro         & RoboTwin  & IL & VLA       & 0.27
  & 0.29 \textcolor{BrickRed}{(+0.02)}
  & 0.48 \textcolor{BrickRed}{(+0.21)} \\
can                        & Qwen3 Coder-Plus       & Robomimic & IL & RNN       & 0.84
  & \textcolor{gray}{\textit{bug}}
  & 0.94 \textcolor{BrickRed}{(+0.10)} \\
square (IL)                & DeepSeek V3.2-Thinking         & Robomimic & IL & RNN       & 0.68
  & \textcolor{gray}{\textit{bug}}
  & 0.84 \textcolor{BrickRed}{(+0.16)} \\
tool-hang                  & Gemini 3.0 Pro         & Robomimic & IL & RNN       & 0.02
  & 0.08 \textcolor{BrickRed}{(+0.06)}
  & 0.08 \textcolor{BrickRed}{(+0.06)} \\
square (RL)                & Gemini 3.0 Pro         & Robomimic & RL & VAE       & 0.12
  & 0.48 \textcolor{BrickRed}{(+0.36)}
  & 0.48 \textcolor{BrickRed}{(+0.36)} \\
push-cube                  & DeepSeek V3.2-Thinking & ManiSkill & RL & MLP       & 0.51
  & 1.00 \textcolor{BrickRed}{(+0.49)}
  & 1.00 \textcolor{BrickRed}{(+0.49)} \\
pick-cube                  & Gemini 3.0 Pro         & ManiSkill & RL & MLP       & 0.92
  & 1.00 \textcolor{BrickRed}{(+0.08)}
  & 1.00 \textcolor{BrickRed}{(+0.08)} \\
peg-insertion-side         & Kimi-K2-Thinking       & ManiSkill & RL & MLP       & 0.00
  & 0.94 \textcolor{BrickRed}{(+0.94)}
  & 0.94 \textcolor{BrickRed}{(+0.94)} \\
pull-cube                  & Gemini 3.0 Pro         & ManiSkill & RL & MLP       & 0.43
  & 1.00 \textcolor{BrickRed}{(+0.57)}
  & 1.00 \textcolor{BrickRed}{(+0.57)} \\
push-t                     & Gemini 3.0 Pro         & ManiSkill & RL & MLP       & 0.73
  & 0.62 \textcolor{OliveGreen}{($-$0.11)}
  & 0.62 \textcolor{OliveGreen}{($-$0.11)} \\
lift-peg-upright           & Kimi-K2-Thinking       & ManiSkill & RL & MLP       & 0.12
  & 1.00 \textcolor{BrickRed}{(+0.88)}
  & 1.00 \textcolor{BrickRed}{(+0.88)} \\
peg-insert-side-mt10$^{a}$ & Gemini 3.0 Pro         & MetaWorld & RL & MLP       & 0.66
  & 0.92 \textcolor{BrickRed}{(+0.26)}
  & 1.00 \textcolor{BrickRed}{(+0.34)} \\
pick-place-mt10            & Gemini 3.0 Pro         & MetaWorld & RL & MLP       & 0.36
  & 0.56 \textcolor{BrickRed}{(+0.20)}
  & 0.78 \textcolor{BrickRed}{(+0.42)} \\
push-mt10                  & Gemini 3.0 Pro         & MetaWorld & RL & MLP       & 0.54
  & 0.34 \textcolor{OliveGreen}{($-$0.20)}
  & 0.92 \textcolor{BrickRed}{(+0.38)} \\
hand-insert-st$^{b}$       & Gemini 3.0 Pro         & MetaWorld & RL & MLP       & 0.50
  & 0.40 \textcolor{OliveGreen}{($-$0.10)}
  & 1.00 \textcolor{BrickRed}{(+0.50)} \\
pick-out-of-hole-st        & Claude Opus 4.5        & MetaWorld & RL & MLP       & 0.50
  & 0.85 \textcolor{BrickRed}{(+0.35)}
  & 1.00 \textcolor{BrickRed}{(+0.50)} \\
coffee-pull-st             & Claude Opus 4.5        & MetaWorld & RL & MLP       & 0.15
  & 0.75 \textcolor{BrickRed}{(+0.60)}
  & 0.95 \textcolor{BrickRed}{(+0.80)} \\
pick-place-st              & Multiple Models        & MetaWorld & RL & MLP       & 0.85
  & 0.70 \textcolor{OliveGreen}{($-$0.15)}
  & 1.00 \textcolor{BrickRed}{(+0.15)} \\
\midrule
\textbf{Average}           & \NA  & \NA & \NA & \NA   & \textbf{0.47}
  & \textbf{0.60} \textcolor{BrickRed}{(+0.13)}
  & \textbf{0.81} \textcolor{BrickRed}{(+0.34)} \\
\bottomrule
\end{tabular}
\end{table}

\begin{figure}[htbp]
    \centering
    \includegraphics[width=\textwidth]{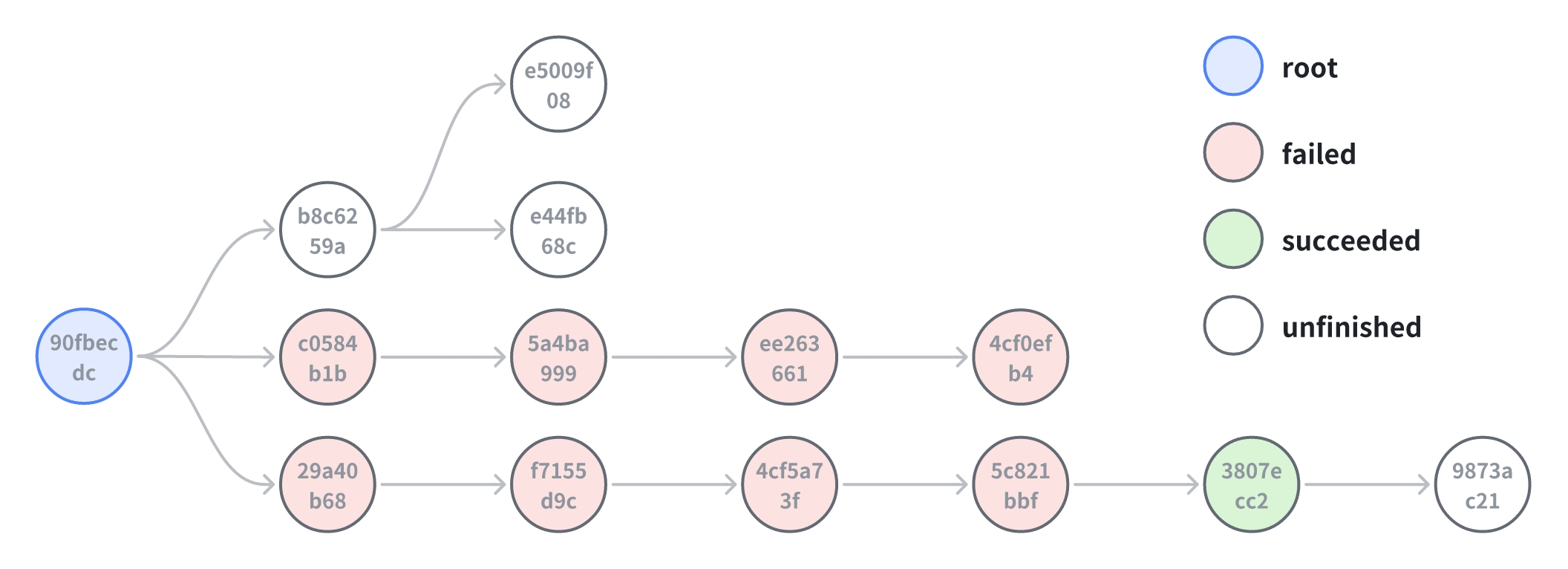}
    \captionsetup{justification=RaggedRight, singlelinecheck=false}
    \caption{\textbf{Extended Data Fig.~1 $|$ Representative feedback-guided search graph.}
    Nodes denote embodied policy-development attempts stored in the tree-structured memory. The root node marks the initial attempt; failed, succeeded, and unfinished nodes indicate the execution outcome of later branches; directed edges show inheritance between attempts. The graph illustrates how \textsc{RoboCoach} retains failed attempts as diagnostic evidence while allowing later rounds to extend the current branch, roll back to a better node, or restart from the baseline according to measured physical outcomes.}
    \label{fig:ed_search_graph}
\end{figure}

\begin{table}[htbp]
\caption{\textbf{Extended Data Table~2 $|$ Benchmark tasks and agent
performance in the from-scratch setting.}
``w/o agentic'': single-shot generation without execution feedback.
``agentic'': iteratively optimised generation via RoboCoach.
\textbf{Aggregation rule:} ``Best LLM'' is selected per task by highest
\textit{agentic} success; averages are task-oracle aggregates, not
model-wise means.
\textbf{Bug handling:} entries marked \textit{bug} are failed runs and
count as 0.00 success when computing averages.
Values: task success rate (fraction of 100 episodes).}
\label{tab:benchmark_performance_scratch}
\centering
\hspace*{\ExtendedDataTableShift}
\setlength{\tabcolsep}{4pt}
\fontfamily{ptm}\selectfont
\setlength{\tabcolsep}{2pt}
\fontsize{7.5pt}{9pt}\selectfont

\begin{tabular}{lllllcll}
\toprule
\textbf{Task} & \textbf{Best LLM} & \textbf{Env.} & \textbf{Mode} &
\textbf{Arch.} & \textbf{Human} &
\textbf{w/o agentic} & \textbf{agentic} \\
\midrule
button-press-topdown-mt10 & Claude Opus 4.5  & MetaWorld & RL & MLP & 1.00
  & 1.00 \textcolor{BrickRed}{(+0.00)}
  & 1.00 \textcolor{BrickRed}{(+0.00)} \\
door-open-mt10            & Multiple Models  & MetaWorld & RL & MLP & 1.00
  & 1.00 \textcolor{BrickRed}{(+0.00)}
  & 1.00 \textcolor{BrickRed}{(+0.00)} \\
drawer-close-mt10         & Multiple Models  & MetaWorld & RL & MLP & 1.00
  & 1.00 \textcolor{BrickRed}{(+0.00)}
  & 1.00 \textcolor{BrickRed}{(+0.00)} \\
drawer-open-mt10          & Multiple Models  & MetaWorld & RL & MLP & 1.00
  & 1.00 \textcolor{BrickRed}{(+0.00)}
  & 1.00 \textcolor{BrickRed}{(+0.00)} \\
reach-mt10                & GPT-5.2          & MetaWorld & RL & MLP & 1.00
  & 0.10 \textcolor{OliveGreen}{($-$0.90)}
  & 1.00 \textcolor{BrickRed}{(+0.00)} \\
window-close-mt10         & Multiple Models  & MetaWorld & RL & MLP & 1.00
  & 1.00 \textcolor{BrickRed}{(+0.00)}
  & 1.00 \textcolor{BrickRed}{(+0.00)} \\
window-open-mt10          & Multiple Models  & MetaWorld & RL & MLP & 1.00
  & 1.00 \textcolor{BrickRed}{(+0.00)}
  & 1.00 \textcolor{BrickRed}{(+0.00)} \\
pick-cube-scratch         & Multiple Models  & ManiSkill & RL & MLP & 0.92
  & 1.00 \textcolor{BrickRed}{(+0.08)}
  & 1.00 \textcolor{BrickRed}{(+0.08)} \\
pull-cube-scratch         & Multiple Models  & ManiSkill & RL & MLP & 0.43
  & 1.00 \textcolor{BrickRed}{(+0.57)}
  & 1.00 \textcolor{BrickRed}{(+0.57)} \\
lift-peg-upright-scratch  & Multiple Models  & ManiSkill & RL & MLP & 0.12
  & 1.00 \textcolor{BrickRed}{(+0.88)}
  & 1.00 \textcolor{BrickRed}{(+0.88)} \\
poke-cube-scratch         & Kimi-K2-Thinking & ManiSkill & RL & MLP & 0.92
  & 0.62 \textcolor{OliveGreen}{($-$0.30)}
  & 0.94 \textcolor{BrickRed}{(+0.02)} \\
\midrule
\textbf{Average}          & \NA  & \NA & \NA & \NA  & \textbf{0.85}
  & \textbf{0.88} \textcolor{BrickRed}{(+0.03)}
  & \textbf{0.99} \textcolor{BrickRed}{(+0.14)} \\
\bottomrule
\end{tabular}
\end{table}

\begin{table}[htbp]
\centering
\caption{\textbf{Extended Data Table~3 $|$ Model-wise benchmark-wide performance.}
Mean task success rates are computed across all 32 benchmark tasks, with failed runs marked \textit{bug} counted as zero success.
The human expert reference is the 32-task platform-reference mean (0.602).}
\label{tab:modelwise_performance}
\fontfamily{ptm}\selectfont
\setlength{\tabcolsep}{4pt}
\fontsize{7.5pt}{9pt}\selectfont
\begin{tabular}{p{0.27\textwidth} c c c c}
\toprule
\textbf{Base model} & \textbf{Non-agentic mean} & \textbf{\textsc{RoboCoach} mean} & \textbf{Uplift} & \textbf{Above human reference} \\
\midrule
Claude Opus~4.5        & 0.577 & 0.785 & +20.8 pp & Yes \\
DeepSeek V3.2-Thinking & 0.515 & 0.778 & +26.2 pp & Yes \\
GLM-4.6                & 0.270 & 0.687 & +41.7 pp & Yes \\
GPT-5.2                & 0.406 & 0.770 & +36.4 pp & Yes \\
Gemini~3.0 Pro         & 0.398 & 0.802 & +40.5 pp & Yes \\
Kimi-K2-Thinking       & 0.418 & 0.754 & +33.7 pp & Yes \\
Qwen3 Coder-Plus       & 0.156 & 0.533 & +37.8 pp & No \\
\midrule
\textbf{Mean}          & \textbf{0.391} & \textbf{0.730} & \textbf{+33.9 pp} & \NA \\
\bottomrule
\end{tabular}
\end{table}

\begin{table}[t]
\centering
\caption{\textbf{Extended Data Table~4 $|$ Ablation on information sources and branching search.}
We report the average performance over 4 tasks in RoboTwin under different feedback and training settings.}
\label{tab:ablation_info_sources}
\small
\setlength{\tabcolsep}{3pt}

\fontfamily{ptm}\selectfont
\setlength{\tabcolsep}{2pt}
\fontsize{7.5pt}{9pt}\selectfont

\hspace*{\ExtendedDataTableShift}\hspace*{-2.8cm}
\begin{tabular}{l l l c c c c c}
\toprule
\textbf{Task} & \textbf{LLM} & \textbf{Sim} & \textbf{Video} & \textbf{Text Signals} & \textbf{Quant. Signals} & \textbf{Branching Search} & \textbf{Performance} \\
\midrule
4 tasks average & Claude Opus 4.5 & RoboTwin & \checkmark & \checkmark & \checkmark & \checkmark & \textbf{0.61} \\
4 tasks average & Claude Opus 4.5 & RoboTwin & \checkmark & \checkmark & \checkmark &  & 0.52 \\
4 tasks average & Claude Opus 4.5 & RoboTwin & \checkmark &  & \checkmark & \checkmark & 0.50 \\
4 tasks average & Claude Opus 4.5 & RoboTwin &  & \checkmark & \checkmark & \checkmark & 0.56 \\
4 tasks average & Claude Opus 4.5 & RoboTwin & \checkmark & \checkmark &  & \checkmark & 0.57 \\
4 tasks average & Claude Opus 4.5 & RoboTwin &  &  &  &  & 0.47 \\
4 tasks average & human & RoboTwin &  &  &  &  & 0.49 \\
\bottomrule
\end{tabular}
\end{table}

\begin{table}[htbp]
\centering
\caption{\textbf{Extended Data Table~5 $|$ Real-robot transfer results.}
Task-level simulation and real-robot success rates are reported under matched budgets for the platform reference and \textsc{RoboCoach}.
The first two tasks use mixed simulation--real finetuning and 10 real-robot evaluation rollouts; the last two use real-only finetuning and 18 real-robot evaluation rollouts.
The stack-cube reference is the matched human-engineered configuration evaluated under the same transfer protocol.}
\label{tab:real_robot_transfer_results}
\fontfamily{ptm}\selectfont
\setlength{\tabcolsep}{2pt}
\fontsize{7pt}{8.2pt}\selectfont
\begin{tabular}{p{0.19\textwidth} p{0.09\textwidth} p{0.20\textwidth} c c c c c}
\toprule
\textbf{Task} & \textbf{Hardware} & \textbf{Finetuning regime} & \textbf{Sim ref.} & \textbf{Sim RC} & \textbf{Real ref.} & \textbf{Real RC} & \textbf{Rollouts} \\
\midrule
banana-to-plate       & Lab~1 & Mixed simulation--real & 0.72 & 0.77 & 0.30 & 0.40 & 10 \\
stack-cube            & Lab~1 & Mixed simulation--real & 0.35 & 0.51 & 0.30 & 0.50 & 10 \\
place-container-plate & Lab~2 & Real-only              & 0.63 & 0.77 & 0.61 & 0.83 & 18 \\
beat-block-hammer     & Lab~2 & Real-only              & 0.41 & 0.49 & 0.83 & 1.00 & 18 \\
\midrule
\textbf{Average}      & \NA  & \NA                    & \textbf{0.53} & \textbf{0.64} & \textbf{0.51} & \textbf{0.68} & \NA \\
\bottomrule
\end{tabular}
\end{table}

\begin{table}[p]
\centering
\caption{\textbf{Extended Data Table~6 $|$ Task abbreviations and benchmark metadata.}
Abbreviations are used in the main figures and Extended Data tables.
Scratch tasks are marked with a trailing asterisk.}
\label{tab:task_abbreviations_metadata}
\fontfamily{ptm}\selectfont
\setlength{\tabcolsep}{3pt}
\fontsize{7pt}{8.2pt}\selectfont
\begin{tabular}{p{0.09\textwidth} p{0.29\textwidth} l l l l}
\toprule
\textbf{Abbrev.} & \textbf{Full task name} & \textbf{Platform} & \textbf{Setting} & \textbf{Mode} & \textbf{Arch.} \\
\midrule
ham    & beat-block-hammer                  & RoboTwin  & Improving & IL & ACT \\
btl    & adjust-bottle                      & RoboTwin  & Improving & IL & ACT \\
cab    & put-object-cabinet                 & RoboTwin  & Improving & IL & Diffusion \\
phn    & place-phone-stand                  & RoboTwin  & Improving & IL & VLA \\
can    & can                                & Robomimic & Improving & IL & RNN \\
sqI    & square (IL)                        & Robomimic & Improving & IL & RNN \\
tool   & tool-hang                          & Robomimic & Improving & IL & RNN \\
sqR    & square (RL)                        & Robomimic & Improving & RL & VAE \\
push   & push-cube                          & ManiSkill & Improving & RL & MLP \\
pick   & pick-cube                          & ManiSkill & Improving & RL & MLP \\
peg    & peg-insertion-side                 & ManiSkill & Improving & RL & MLP \\
pull   & pull-cube                          & ManiSkill & Improving & RL & MLP \\
pT     & push-t                             & ManiSkill & Improving & RL & MLP \\
lift   & lift-peg-upright                   & ManiSkill & Improving & RL & MLP \\
pegM   & peg-insert-side-mt10               & MetaWorld & Improving & RL & MLP \\
ppM    & pick-place-mt10                    & MetaWorld & Improving & RL & MLP \\
psM    & push-mt10                          & MetaWorld & Improving & RL & MLP \\
hand   & hand-insert-st                     & MetaWorld & Improving & RL & MLP \\
hole   & pick-out-of-hole-st                & MetaWorld & Improving & RL & MLP \\
cof    & coffee-pull-st                     & MetaWorld & Improving & RL & MLP \\
ppS    & pick-place-st                      & MetaWorld & Improving & RL & MLP \\
press* & button-press-topdown-mt10-scratch  & MetaWorld & Scratch   & RL & MLP \\
door*  & door-open-mt10-scratch             & MetaWorld & Scratch   & RL & MLP \\
drC*   & drawer-close-mt10-scratch          & MetaWorld & Scratch   & RL & MLP \\
drO*   & drawer-open-mt10-scratch           & MetaWorld & Scratch   & RL & MLP \\
reach* & reach-mt10-scratch                 & MetaWorld & Scratch   & RL & MLP \\
winC*  & window-close-mt10-scratch          & MetaWorld & Scratch   & RL & MLP \\
winO*  & window-open-mt10-scratch           & MetaWorld & Scratch   & RL & MLP \\
pick*  & pick-cube-scratch                  & ManiSkill & Scratch   & RL & MLP \\
pull*  & pull-cube-scratch                  & ManiSkill & Scratch   & RL & MLP \\
lift*  & lift-peg-upright-scratch           & ManiSkill & Scratch   & RL & MLP \\
poke*  & poke-cube-scratch                  & ManiSkill & Scratch   & RL & MLP \\
\bottomrule
\end{tabular}
\end{table}

\begin{table}[t]
\centering
\caption{\textbf{Extended Data Table~7 $|$ Harness comparison on RoboTwin IL.}
Mean success rate over 45 RoboTwin imitation-learning tasks using GPT-5.2 as the base model.}
\label{tab:harness_comparison_robotwin_il_gpt52}
\small
\setlength{\tabcolsep}{8pt}
\begin{tabular}{l c c c}
\toprule
\textbf{Harness} & \textbf{Base LLM} & \textbf{Benchmark subset} & \textbf{Mean success} \\
\midrule
\textsc{RoboCoach} & GPT-5.2 & RoboTwin IL, 45 tasks & 0.59 \\
OpenClaw & GPT-5.2 & RoboTwin IL, 45 tasks & 0.10 \\
Hermes & GPT-5.2 & RoboTwin IL, 45 tasks & 0.08 \\
ARIS & GPT-5.2 & RoboTwin IL, 45 tasks & 0.38\\
\bottomrule
\end{tabular}
\end{table}

% Appendix migrated from RoboCoach_0611_SAI/appendix.tex
\clearpage
\section*{Appendix}
\setcounter{figure}{0}
\setcounter{table}{0}
\renewcommand{\thefigure}{A\arabic{figure}}
\renewcommand{\thetable}{A\arabic{table}}
\renewcommand{\theHfigure}{appendix.\arabic{figure}}
\renewcommand{\theHtable}{appendix.\arabic{table}}
\captionsetup[figure]{labelformat=default}
\captionsetup[table]{labelformat=default}

\subsection{Task Description}
\label{sec:task_desc}
\subsubsection{RoboTwin}
In this platform, the participating robot is a dual-arm Aloha-AgileX. We report the mean success rate of 100 episodes.        

\noindent
\textbf{Beat-block-hammer.}
There is a hammer and a block on the table, use the arm to grab the hammer and beat the block. 

\noindent
\textbf{Adjust-bottle.}
Pick up the bottle on the table headup with the correct arm. 

\noindent
\textbf{put-object-cabinet.} 
Use one arm to open the cabinet's drawer, and use the other arm to put the object on the table to the drawer. 

\noindent
\textbf{Place-phone-stand.}
Pick up the phone and put it on the phone stand.  

\subsubsection{Robomimic}

\noindent
\textbf{Can.} In the Can task, a Franka Emika Panda robot must grasp a coke can located randomly within a large bin and precisely place it into a smaller specific target bin. This task introduces higher complexity than simple lifting by requiring object transport and placement, with the primary evaluation metric being the success rate of the can landing in the target bin.

\noindent
\textbf{Square.} The Square task is a high-precision simulated environment where a Franka Emika Panda robot must pick up a hollow square nut and thread it onto a stationary matching square rod. This task tests the robot's ability to perform precise alignments and insertions, and the model is evaluated based on the success rate of fully seating the nut onto the rod.

\noindent
\textbf{Tool-hang.} Tool Hang is considered the most difficult task in the suite, requiring a Franka Emika Panda robot (in simulation or real-world) to first assemble a frame by inserting a hook piece into a base and then hang a wrench onto the newly assembled hook. This task evaluates the robot's capacity for dexterous, rotation-heavy manipulation across multiple stages, using the task completion success rate as the metric.

\subsubsection{ManiSkill}

\noindent
\textbf{Push-cube.} In the Push-Cube task, the objective is for a robot (typically a Fetch mobile manipulator or a Franka Emika Panda arm) to push a cube across a surface to a specified target goal region. This task focuses on non-prehensile manipulation and precise control of object dynamics without grasping, and performance is evaluated based on the success rate of the cube overlapping with the target area within a specific tolerance.

\noindent
\textbf{Pick-cube.}
The Pick-Cube task serves as a fundamental grasp-and-lift benchmark where a Franka Emika Panda robot arm must locate a specific cube on a tabletop, grasp it securely, and lift it to a target height. This task tests the agent's inverse kinematics and grasping stability, with the primary metric being the success rate of the object remaining elevated above a height threshold at the end of the episode.

\noindent
\textbf{Peg-insertion-side.}
Peg-Insertion-Side is a high-precision assembly task requiring a Franka Emika Panda robot to grasp a peg and insert it into a hole that is oriented horizontally (sideways) on a box. This environment challenges the robot's alignment accuracy and ability to handle contact forces during insertion, measuring performance by the success rate of the peg penetrating the hole to a required depth.

\noindent
\textbf{Pull-cube.}
In the Pull-Cube task, the agent controls a robot (often a Franka Emika Panda or Fetch) to retrieve a cube and move it towards a target location, typically requiring a pulling motion rather than a push. This task emphasizes the ability to manipulate objects to bring them closer to a specific zone or the robot base, evaluated by the success rate of the cube reaching the designated target coordinates.

\noindent
\textbf{Push-t.}
The Push-T task involves a robot (typically a Franka Emika Panda or a simplified End-Effector agent) pushing a T-shaped block to strictly align it with a corresponding T-shaped target outline on the table. Due to the T-block's non-convex geometry, the agent must execute complex multi-stage pushing actions to correct both position and orientation, with the metric being the success rate based on the overlap (IoU) between the block and the target.

\noindent
\textbf{Lift-peg-upright.}
Lift-Peg-Upright introduces a reorientation challenge where a Franka Emika Panda robot must pick up a peg that is initially lying flat (horizontal) on the table, reorient it to a vertical upright position, and lift it. This task tests dexterous manipulation capabilities involving potential regasping or pivot-lifting strategies, and is evaluated by the success rate of the peg achieving the correct vertical orientation while being lifted to the target height.

\subsubsection{MetaWorld}

In this platform, the participating robot is Rethink Robotics Sawyer. We present three experimental configurations: mt10, mt10-scratch, and single-task (st). Both mt10 and mt10-scratch adhere to the standard MetaWorld-mt10 training protocol, whereas the single-task setting restricts training and evaluation to isolated task environments. Moreover, mt10 is an 'improve' setting, and mt10-scratch challenges the agent to develop the implementation from a sparse codebase. To ensure a unified evaluation protocol, we report the average success rate over 100 episodes for each task across all three settings. The related tasks are described below: 

\noindent
\textbf{peg-insert-side-mt10.} 
Insert a peg sideways. Randomize peg and goal positions.

\noindent
\textbf{pick-place-mt10.}
Pick and place a puck to a goal. Randomize puck and goal positions.

\noindent
\textbf{push-mt10.}
Push the puck to a goal. Randomize puck and goal positions.

\noindent
\textbf{button-press-topdown-mt10-scratch.}
Press a button from the top. Randomize button positions

\noindent
\textbf{door-open-mt10-scratch.}
Open a door with a revolving joint. Randomize door positions.

\noindent
\textbf{drawer-close-mt10-scratch.}
Push and close a drawer. Randomize the drawer positions.

\noindent
\textbf{drawer-open-mt10-scratch.}
Open a drawer. Randomize drawer positions.

\noindent
\textbf{reach-mt10-scratch.}
Reach a goal position. Randomize the goal positions.

\noindent
\textbf{window-close-mt10-scratch.}
Push and close a window. Randomize window positions.

\noindent
\textbf{window-open-mt10-scratch.}
Push and open a window. Randomize window positions.

\noindent
\textbf{hand-insert-st.}
Insert the gripper into a hole.

\noindent
\textbf{pick-out-of-hole-st.}
Pick up a puck from a hole. Randomize puck and goal positions.

\noindent
\textbf{coffee-pull-st.}
Grasp a mug and pull it across the table to a designated target location.

\noindent
\textbf{pick-place-st.}
Pick and place a puck to a goal. Randomize puck and goal positions.

\subsection{Real-robot system setup}
\label{sec:real_robot_setup}

Real-robot transfer experiments are conducted with Franka Research~3 (FR3) 7-DoF robot arms equipped with the default Franka Hand two-finger gripper.
Perception uses a single Intel RealSense D435if camera.
Raw camera streams are resized and cropped to $224 \times 224$ pixels before being passed to the policy network.
Robot control is implemented through the \texttt{franky} library, a high-level Python/C++ interface built on top of \texttt{libfranka}.
All data collection and policy execution run at 10~Hz, with both the robot state and action space represented as 7-dimensional joint positions.

The four real-robot transfer tasks are distributed across two hardware laboratories.
The first two tasks reported in Fig.~\ref{fig:sim2sim2real}, banana-to-plate and stack-cube, are run in Lab~1.
The remaining two tasks, place-container-plate and beat-block-hammer, are run in Lab~2.

\subsection{Complete RoboCoach-Bench performances}

Tab.~\ref{tab:embocoach-complete} presents a comprehensive comparison of multiple LLM agents across all tasks in RoboCoach-Bench, reporting their performance in terms of success rates. The evaluated LLMs include Claude-Opus-4.5, Gemini-3.0-Pro, Gemini-3-pro, GPT-5.2, Kimi-K2-Thinking, Qwen3-Coder-Plus, GLM-4.6, and DeepSeek-V3.2-Thinking. In addition, we report human performance obtained by directly training policies using the original embodied codebases. The experiments span four representative robotic simulation environments: RoboTwin, RoboMimic, ManiSkill, and MetaWorld.
We compare two LLM coding paradigms: one-shot planning, where the LLM generates a single plan and executes the embodied experiment once, and iterative planning, which incorporates Monte Carlo Tree Search (MCTS) for global solution searching based on execution feedback.

{\fontsize{7.5}{8.5}\selectfont % 或 \footnotesize 先试，放不下再 \scriptsize
\setlength{\tabcolsep}{1.0pt}   % 3pt/2pt/1.5pt 逐步调小
\renewcommand{\arraystretch}{1.05}

\begin{xltabular}{\textwidth}{c c c c c c l l}
\caption{Complete performance comparison on RoboCoach-Bench.}
\label{tab:embo_xl} \\
\toprule
\multirow{2}{*}{\textbf{Embodied Task}} &
\multirow{2}{*}{\textbf{LLM}} &
\multirow{2}{*}{\textbf{Environment}} &
\multirow{2}{*}{\makecell{\textbf{Learning}\\\textbf{Mode}}} &
\multirow{2}{*}{\makecell{\textbf{Embodied}\\\textbf{Architecture}}} &
\multirow{2}{*}{\makecell{\textbf{Human}\\\textbf{Perf.}}} &
        \multicolumn{2}{c}{\textbf{LLM Perf.}} \\
        & & & & & & \textbf{w/o Agentic} & \textbf{Agentic} \\
\midrule
\endfirsthead

\toprule
\multirow{2}{*}{\textbf{Embodied Task}} &
\multirow{2}{*}{\textbf{LLM of Agent}} &
\multirow{2}{*}{\textbf{Environment}} &
\multirow{2}{*}{\makecell{\textbf{Learning}\\\textbf{Mode}}} &
\multirow{2}{*}{\makecell{\textbf{Embodied}\\\textbf{Architecture}}} &
\multirow{2}{*}{\makecell{\textbf{Human}\\\textbf{Perf.}}} &
\multicolumn{2}{c}{\textbf{LLM Agent Perf.}} \\
& & & & & & \textbf{w/o Agentic} & \textbf{Agentic} \\
\midrule
\endhead

\midrule
\multicolumn{8}{r}{\small Continued on next page} \\
\endfoot

\bottomrule
\endlastfoot

% ---- 在这里开始填内容 ----
% task & llm & env & mode & arch & human & wo & agentic \\
% ...
beat-block-hammer                                  & Claude-Opus-4.5                                            & RoboTwin                     & IL                                                          & ACT                                                                 & 0.40                                                & 0.40 \textcolor{BrickRed}{(+0.00 $\uparrow$)}     & 0.54 \textcolor{BrickRed}{(+0.14 $\uparrow$)}     \\
beat-block-hammer                                  & Gemini-3.0-Pro                                             & RoboTwin                     & IL                                                          & ACT                                                                 & 0.40                                                & 0.00 \textcolor{OliveGreen}{(-0.40 $\downarrow$)} & 0.50 \textcolor{BrickRed}{(+0.10 $\uparrow$)}     \\
beat-block-hammer                                  & GPT-5.2                                                   & RoboTwin                     & IL                                                          & ACT                                                                 & 0.40                                                & 0.32 \textcolor{OliveGreen}{(-0.08 $\downarrow$)} & 0.48 \textcolor{BrickRed}{(+0.08 $\uparrow$)}     \\
beat-block-hammer                                  & Kimi-K2-Thinking                                           & RoboTwin                     & IL                                                          & ACT                                                                 & 0.40                                                & 0.35 \textcolor{OliveGreen}{(-0.05 $\downarrow$)} & 0.53 \textcolor{BrickRed}{(+0.13 $\uparrow$)}     \\
beat-block-hammer                                  & qwen3-coder-plus                                           & RoboTwin                     & IL                                                          & ACT                                                                 & 0.40                                                & 0.36 \textcolor{OliveGreen}{(-0.04 $\downarrow$)} & 0.50 \textcolor{BrickRed}{(+0.10 $\uparrow$)}     \\
beat-block-hammer                                  & GLM-4.6                                                   & RoboTwin                     & IL                                                          & ACT                                                                 & 0.40                                                & 0.33 \textcolor{OliveGreen}{(-0.07 $\downarrow$)} & 0.42 \textcolor{BrickRed}{(+0.02 $\uparrow$)}     \\
beat-block-hammer                                  & DeepSeek-V3.2-Thinking                                     & RoboTwin                     & IL                                                          & ACT                                                                 & 0.40                                                & 0.28 \textcolor{OliveGreen}{(-0.12 $\downarrow$)} & 0.46 \textcolor{BrickRed}{(+0.06 $\uparrow$)}     \\
adjust-bottle                                      & Claude-Opus-4.5                                            & RoboTwin                     & IL                                                          & ACT                                                                 & 0.92                                                & 0.92 \textcolor{BrickRed}{(+0.00 $\uparrow$)}     & 0.95 \textcolor{BrickRed}{(+0.03 $\uparrow$)}     \\
adjust-bottle                                      & Gemini-3.0-Pro                                             & RoboTwin                     & IL                                                          & ACT                                                                 & 0.92                                                & 0.94 \textcolor{BrickRed}{(+0.02 $\uparrow$)}     & 0.98 \textcolor{BrickRed}{(+0.06 $\uparrow$)}     \\
adjust-bottle                                      & GPT-5.2                                                   & RoboTwin                     & IL                                                          & ACT                                                                 & 0.92                                                & 0.00 \textcolor{OliveGreen}{(-0.92 $\downarrow$)} & 0.93 \textcolor{BrickRed}{(+0.01 $\uparrow$)}     \\
adjust-bottle                                      & Kimi-K2-Thinking                                           & RoboTwin                     & IL                                                          & ACT                                                                 & 0.92                                                & 0.92 \textcolor{BrickRed}{(+0.00 $\uparrow$)}     & 0.96 \textcolor{BrickRed}{(+0.04 $\uparrow$)}     \\
adjust-bottle                                      & qwen3-coder-plus                                           & RoboTwin                     & IL                                                          & ACT                                                                 & 0.92                                                & 0.94 \textcolor{BrickRed}{(+0.02 $\uparrow$)}     & 0.96 \textcolor{BrickRed}{(+0.04 $\uparrow$)}     \\
adjust-bottle                                      & GLM-4.6                                                   & RoboTwin                     & IL                                                          & ACT                                                                 & 0.92                                                & 0.85 \textcolor{OliveGreen}{(-0.07 $\downarrow$)} & 0.94 \textcolor{BrickRed}{(+0.02 $\uparrow$)}     \\
adjust-bottle                                      & DeepSeek-V3.2-Thinking                                     & RoboTwin                     & IL                                                          & ACT                                                                 & 0.92                                                & 0.89 \textcolor{OliveGreen}{(-0.03 $\downarrow$)} & 0.93 \textcolor{BrickRed}{(+0.01 $\uparrow$)}     \\
put-object-cabinet                                 & Claude-Opus-4.5                                            & RoboTwin                     & IL                                                          & Diffusion                                                           & 0.36                                                & 0.33 \textcolor{OliveGreen}{(-0.03 $\downarrow$)} & 0.48 \textcolor{BrickRed}{(+0.12 $\uparrow$)}     \\
put-object-cabinet                                 & Gemini-3.0-Pro                                             & RoboTwin                     & IL                                                          & Diffusion                                                           & 0.36                                                & 0.36 \textcolor{BrickRed}{(+0.00 $\uparrow$)}     & 0.46 \textcolor{BrickRed}{(+0.10 $\uparrow$)}     \\
put-object-cabinet                                 & GPT-5.2                                                   & RoboTwin                     & IL                                                          & Diffusion                                                           & 0.36                                                & 0.35 \textcolor{OliveGreen}{(-0.01 $\downarrow$)} & 0.42 \textcolor{BrickRed}{(+0.06 $\uparrow$)}     \\
put-object-cabinet                                 & Kimi-K2-Thinking                                           & RoboTwin                     & IL                                                          & Diffusion                                                           & 0.36                                                & 0.34 \textcolor{OliveGreen}{(-0.02 $\downarrow$)} & 0.54 \textcolor{BrickRed}{(+0.18 $\uparrow$)}     \\
put-object-cabinet                                 & qwen3-coder-plus                                           & RoboTwin                     & IL                                                          & Diffusion                                                           & 0.36                                                & 0.40 \textcolor{BrickRed}{(+0.04 $\uparrow$)}     & 0.50 \textcolor{BrickRed}{(+0.14 $\uparrow$)}     \\
put-object-cabinet                                 & GLM-4.6                                                   & RoboTwin                     & IL                                                          & Diffusion                                                           & 0.36                                                & 0.33 \textcolor{OliveGreen}{(-0.03 $\downarrow$)} & 0.34 \textcolor{OliveGreen}{(-0.02 $\downarrow$)} \\
put-object-cabinet                                 & DeepSeek-V3.2-Thinking                                     & RoboTwin                     & IL                                                          & Diffusion                                                           & 0.36                                                & 0.38 \textcolor{BrickRed}{(+0.02 $\uparrow$)}     & 0.40 \textcolor{BrickRed}{(+0.04 $\uparrow$)}     \\
place-phone-stand                                  & Claude-Opus-4.5                                            & RoboTwin                     & IL                                                          & VLA                                                                 & 0.27                                                & 0.23 \textcolor{OliveGreen}{(-0.04 $\downarrow$)} & 0.46 \textcolor{BrickRed}{(+0.19 $\uparrow$)}     \\
place-phone-stand                                  & Gemini-3.0-Pro                                             & RoboTwin                     & IL                                                          & VLA                                                                 & 0.27                                                & 0.29 \textcolor{BrickRed}{(+0.02 $\uparrow$)}     & 0.48 \textcolor{BrickRed}{(+0.21 $\uparrow$)}     \\
place-phone-stand                                  & GPT-5.2                                                   & RoboTwin                     & IL                                                          & VLA                                                                 & 0.27                                                & 0.25 \textcolor{OliveGreen}{(-0.02 $\downarrow$)} & 0.44 \textcolor{BrickRed}{(+0.17 $\uparrow$)}     \\
place-phone-stand                                  & Kimi-K2-Thinking                                           & RoboTwin                     & IL                                                          & VLA                                                                 & 0.27                                                & 0.26 \textcolor{OliveGreen}{(-0.01 $\downarrow$)} & 0.40 \textcolor{BrickRed}{(+0.13 $\uparrow$)}     \\
place-phone-stand                                  & qwen3-coder-plus                                           & RoboTwin                     & IL                                                          & VLA                                                                 & 0.27                                                & 0.28 \textcolor{BrickRed}{(+0.01 $\uparrow$)}     & 0.42 \textcolor{BrickRed}{(+0.15 $\uparrow$)}     \\
place-phone-stand                                  & GLM-4.6                                                   & RoboTwin                     & IL                                                          & VLA                                                                 & 0.27                                                & 0.31 \textcolor{BrickRed}{(+0.04 $\uparrow$)}     & 0.32 \textcolor{BrickRed}{(+0.05 $\uparrow$)}     \\
place-phone-stand                                  & DeepSeek-V3.2-Thinking                                     & RoboTwin                     & IL                                                          & VLA                                                                 & 0.27                                                & 0.26 \textcolor{OliveGreen}{(-0.01 $\downarrow$)} & 0.38 \textcolor{BrickRed}{(+0.11 $\uparrow$)}     \\ 
can                                                & Kimi-K2-Thinking                                           & Robomimic                    & IL                                                          & RNN                                                                 & 0.84                                                & 0.90 \textcolor{BrickRed}{(+0.06 $\uparrow$)}     & 0.90 \textcolor{BrickRed}{(+0.06 $\uparrow$)}     \\
can                                                & DeepSeek-V3.2                                              & Robomimic                    & IL                                                          & RNN                                                                 & 0.84                                                & 0.88 \textcolor{BrickRed}{(+0.04 $\uparrow$)}     & 0.88 \textcolor{BrickRed}{(+0.04 $\uparrow$)}     \\
can                                                & GLM-4.6                                                   & Robomimic                    & IL                                                          & RNN                                                                 & 0.84                                                & 0.82 \textcolor{OliveGreen}{(-0.02 $\downarrow$)} & 0.86 \textcolor{BrickRed}{(+0.02 $\uparrow$)}     \\
can                                                & Gemini-3.0-pro                                             & Robomimic                    & IL                                                          & RNN                                                                 & 0.84                                                & 0.80 \textcolor{OliveGreen}{(-0.04 $\downarrow$)} & 0.84 \textcolor{BrickRed}{(+0.00 $\uparrow$)}     \\
can                                                & GPT-5.2                                        & Robomimic                    & IL                                                          & RNN                                                                 & 0.84                                                & 0.88 \textcolor{BrickRed}{(+0.04 $\uparrow$)}     & 0.88 \textcolor{BrickRed}{(+0.04 $\uparrow$)}     \\
can                                                & Claude-Opus-4.5                                            & Robomimic                    & IL                                                          & RNN                                                                 & 0.84                                                & 0.80 \textcolor{OliveGreen}{(-0.04 $\downarrow$)} & 0.80 \textcolor{OliveGreen}{(-0.04 $\downarrow$)} \\
can                                                & qwen3-coder-plus                                            & Robomimic                    & IL                                                          & RNN                                                                 & 0.84                                                & \textcolor{gray}{bug} & 0.94 \textcolor{BrickRed}{(+0.10 $\uparrow$)} \\
square                                             & Kimi-K2-Thinking                                           & Robomimic                    & IL                                                          & RNN                                                                 & 0.68                                                & 0.70 \textcolor{BrickRed}{(+0.02 $\uparrow$)}     & 0.70 \textcolor{BrickRed}{(+0.02 $\uparrow$)}     \\
square                                             & DeepSeek-V3.2                                              & Robomimic                    & IL                                                          & RNN                                                                 & 0.68                                                & \textcolor{gray}{bug}                                                                               & 0.84 \textcolor{BrickRed}{(+0.16 $\uparrow$)}     \\
square                                             & GLM-4.6                                                   & Robomimic                    & IL                                                          & RNN                                                                 & 0.68                                                & 0.70 \textcolor{BrickRed}{(+0.02 $\uparrow$)}     & 0.70 \textcolor{BrickRed}{(+0.02 $\uparrow$)}     \\
square                                             & Gemini-3.0-pro                                             & Robomimic                    & IL                                                          & RNN                                                                 & 0.68                                                & \textcolor{gray}{bug}                                                                               & 0.66 \textcolor{OliveGreen}{(-0.02 $\downarrow$)} \\
square                                             & GPT-5.2                                        & Robomimic                    & IL                                                          & RNN                                                                 & 0.68                                                & 0.66 \textcolor{OliveGreen}{(-0.02 $\downarrow$)} & 0.66 \textcolor{OliveGreen}{(-0.02 $\downarrow$)} \\
square                                             & Claude-Opus-4.5                                            & Robomimic                    & IL                                                          & RNN                                                                 & 0.68                                                & \textcolor{gray}{bug}                                                                               & 0.70 \textcolor{BrickRed}{(+0.02 $\uparrow$)}     \\
square                                                & qwen3-coder-plus                                            & Robomimic                    & IL                                                          & RNN                                                                 & 0.68                                                & \textcolor{gray}{bug} & 0.70 \textcolor{BrickRed}{(+0.02 $\uparrow$)} \\
tool-hang                                          & Kimi-K2-Thinking                                           & Robomimic                    & IL                                                          & RNN                                                                 & 0.02                                                & 0.02 \textcolor{BrickRed}{(+0.00 $\uparrow$)}     & 0.08 \textcolor{BrickRed}{(+0.06 $\uparrow$)}     \\
tool-hang                                          & DeepSeek-V3.2                                              & Robomimic                    & IL                                                          & RNN                                                                 & 0.02                                                & \textcolor{gray}{bug}                                                                               & 0.08 \textcolor{BrickRed}{(+0.06 $\uparrow$)}     \\
tool-hang                                          & GLM-4.6                                                   & Robomimic                    & IL                                                          & RNN                                                                 & 0.02                                                & \textcolor{gray}{bug}                                                                               & 0.04 \textcolor{BrickRed}{(+0.02 $\uparrow$)}     \\
tool-hang                                          & Gemini-3.0-pro                                             & Robomimic                    & IL                                                          & RNN                                                                 & 0.02                                                & 0.08 \textcolor{BrickRed}{(+0.06 $\uparrow$)}     & 0.08 \textcolor{BrickRed}{(+0.06 $\uparrow$)}     \\
tool-hang                                          & GPT-5.2                                        & Robomimic                    & IL                                                          & RNN                                                                 & 0.02                                                & 0.04 \textcolor{BrickRed}{(+0.02 $\uparrow$)}     & 0.04 \textcolor{BrickRed}{(+0.02 $\uparrow$)}     \\
tool-hang                                          & Claude-Opus-4.5                                            & Robomimic                    & IL                                                          & RNN                                                                 & 0.02                                                & \textcolor{gray}{bug}                                                                               & 0.02 \textcolor{BrickRed}{(+0.00 $\uparrow$)}     \\
tool-hang                                          & qwen3-coder-plus                                            & Robomimic                    & IL                                                          & RNN                                                                 & 0.02                                                & \textcolor{gray}{bug}                                                                               & 0.00 \textcolor{OliveGreen}{(-0.02 $\downarrow$)}     \\
square                                             & Kimi-K2-Thinking                                           & Robomimic                    & RL                                                          & VAE                                                                 & 0.12                                                & 0.22 \textcolor{BrickRed}{(+0.10 $\uparrow$)}     & 0.22 \textcolor{BrickRed}{(+0.10 $\uparrow$)}     \\
square                                             & DeepSeek-V3.2                                              & Robomimic                    & RL                                                          & VAE                                                                 & 0.12                                                & \textcolor{gray}{bug}                                                                               & 0.22 \textcolor{BrickRed}{(+0.10 $\uparrow$)}     \\
square                                             & GLM-4.6                                                   & Robomimic                    & RL                                                          & VAE                                                                 & 0.12                                                & 0.24 \textcolor{BrickRed}{(+0.12 $\uparrow$)}     & 0.24 \textcolor{BrickRed}{(+0.12 $\uparrow$)}     \\
square                                             & Gemini-3.0-pro                                             & Robomimic                    & RL                                                          & VAE                                                                 & 0.12                                                & 0.48 \textcolor{BrickRed}{(+0.36 $\uparrow$)}     & 0.48 \textcolor{BrickRed}{(+0.36 $\uparrow$)}     \\
square                                             & GPT-5.2                                        & Robomimic                    & RL                                                          & VAE                                                                 & 0.12                                                & 0.36 \textcolor{BrickRed}{(+0.24 $\uparrow$)}     & 0.36 \textcolor{BrickRed}{(+0.24 $\uparrow$)}     \\
square                                             & Claude-Opus-4.5                                            & Robomimic                    & RL                                                          & VAE                                                                 & 0.12                                                & 0.46 \textcolor{BrickRed}{(+0.34 $\uparrow$)}     & 0.46 \textcolor{BrickRed}{(+0.34 $\uparrow$)}     \\
square                                             & qwen3-coder-plus                                            & Robomimic                    & RL                                                          & VAE                                                                 & 0.12                                                & \textcolor{gray}{bug}     & 0.46 \textcolor{BrickRed}{(+0.34 $\uparrow$)}     \\
push-cube                                          & Claude-opus-4-5                                            & ManiSkill                    & RL                                                          & MLP                                                                 & 0.51                                                & 1.00 \textcolor{BrickRed}{(+0.49 $\uparrow$)}     & 1.00 \textcolor{BrickRed}{(+0.49 $\uparrow$)}     \\
push-cube                                          & GPT-5.2                                             & ManiSkill                    & RL                                                          & MLP                                                                 & 0.51                                                & 1.00 \textcolor{BrickRed}{(+0.49 $\uparrow$)}     & 1.00 \textcolor{BrickRed}{(+0.49 $\uparrow$)}     \\
push-cube                                          & Gemini-3-pro                              & ManiSkill                    & RL                                                          & MLP                                                                 & 0.51                                                & 1.00 \textcolor{BrickRed}{(+0.49 $\uparrow$)}     & 1.00 \textcolor{BrickRed}{(+0.49 $\uparrow$)}     \\
push-cube                                          & DeepSeek-V3.2-Thinking                                     & ManiSkill                    & RL                                                          & MLP                                                                 & 0.51                                                & 1.00 \textcolor{BrickRed}{(+0.49 $\uparrow$)}     & 1.00 \textcolor{BrickRed}{(+0.49 $\uparrow$)}     \\
push-cube                                          & GLM-4.6                                           & ManiSkill                    & RL                                                          & MLP                                                                 & 0.51                                                & 1.00 \textcolor{BrickRed}{(+0.49 $\uparrow$)}     & 1.00 \textcolor{BrickRed}{(+0.49 $\uparrow$)}     \\
push-cube                                          & Kimi-K2-Thinking                                   & ManiSkill                    & RL                                                          & MLP                                                                 & 0.51                                                & 1.00 \textcolor{BrickRed}{(+0.49 $\uparrow$)}     & 1.00 \textcolor{BrickRed}{(+0.49 $\uparrow$)}     \\
push-cube                                          & qwen3-coder-plus                                   & ManiSkill                    & RL                                                          & MLP                                                                 & 0.51                                                & 1.00 \textcolor{BrickRed}{(+0.49 $\uparrow$)}     & 1.00 \textcolor{BrickRed}{(+0.49 $\uparrow$)}     \\
pick-cube                                          & Claude-opus-4-5                                            & ManiSkill                    & RL                                                          & MLP                                                                 & 0.92                                                & 0.06 \textcolor{OliveGreen}{(-0.86 $\downarrow$)} & 1.00 \textcolor{BrickRed}{(+0.08 $\uparrow$)}     \\
pick-cube                                          & GPT-5.2                                             & ManiSkill                    & RL                                                          & MLP                                                                 & 0.92                                                & 1.00 \textcolor{BrickRed}{(+0.08 $\uparrow$)}     & 1.00 \textcolor{BrickRed}{(+0.08 $\uparrow$)}     \\
pick-cube                                          & Gemini-3-pro                              & ManiSkill                    & RL                                                          & MLP                                                                 & 0.92                                                & 1.00 \textcolor{BrickRed}{(+0.08 $\uparrow$)}     & 1.00 \textcolor{BrickRed}{(+0.08 $\uparrow$)}     \\
pick-cube                                          & DeepSeek-V3.2-Thinking                                     & ManiSkill                    & RL                                                          & MLP                                                                 & 0.92                                                & 0.00 \textcolor{OliveGreen}{(-0.92 $\downarrow$)} & 1.00 \textcolor{BrickRed}{(+0.08 $\uparrow$)}     \\
pick-cube                                          & GLM-4.6                                           & ManiSkill                    & RL                                                          & MLP                                                                 & 0.92                                                & 0.00 \textcolor{OliveGreen}{(-0.92 $\downarrow$)} & 0.12 \textcolor{OliveGreen}{(-0.80 $\downarrow$)} \\
pick-cube                                          & Kimi-K2-Thinking                                   & ManiSkill                    & RL                                                          & MLP                                                                 & 0.92                                                & 1.00 \textcolor{BrickRed}{(+0.08 $\uparrow$)}     & 1.00 \textcolor{BrickRed}{(+0.08 $\uparrow$)}     \\
pick-cube                                          & qwen3-coder-plus                                           & ManiSkill                    & RL                                                          & MLP                                                                 & 0.92                                                & \textcolor{gray}{bug}                                                                               & 1.00 \textcolor{BrickRed}{(+0.08 $\uparrow$)}     \\
peg-insertion-side                                 & Claude-opus-4-5                                            & ManiSkill                    & RL                                                          & MLP                                                                 & 0.00                                                & 0.62 \textcolor{BrickRed}{(+0.62 $\uparrow$)}     & 0.88 \textcolor{BrickRed}{(+0.88 $\uparrow$)}     \\
peg-insertion-side                                 & Gemini-3-pro                              & ManiSkill                    & RL                                                          & MLP                                                                 & 0.00                                                & 0.31 \textcolor{BrickRed}{(+0.31 $\uparrow$)}     & 0.31 \textcolor{BrickRed}{(+0.31 $\uparrow$)}     \\
peg-insertion-side                                 & GLM-4.6                                           & ManiSkill                    & RL                                                          & MLP                                                                 & 0.00                                                & 0.06 \textcolor{BrickRed}{(+0.06 $\uparrow$)}     & 0.81 \textcolor{BrickRed}{(+0.81 $\uparrow$)}     \\
peg-insertion-side                                 & Kimi-K2-Thinking                                   & ManiSkill                    & RL                                                          & MLP                                                                 & 0.00                                                & 0.94 \textcolor{BrickRed}{(+0.94 $\uparrow$)}     & 0.94 \textcolor{BrickRed}{(+0.94 $\uparrow$)}     \\
peg-insertion-side                                 & GPT-5.2                                   & ManiSkill                    & RL                                                          & MLP                                                                 & 0.00                                                & 0.62 \textcolor{BrickRed}{(+0.62 $\uparrow$)}     & 0.75 \textcolor{BrickRed}{(+0.75 $\uparrow$)}     \\
peg-insertion-side                                 & DeepSeek-V3.2-Thinking                                   & ManiSkill                    & RL                                                          & MLP                                                                 & 0.00                                                & 0.81 \textcolor{BrickRed}{(+0.81 $\uparrow$)}     & 0.81 \textcolor{BrickRed}{(+0.81 $\uparrow$)}     \\
peg-insertion-side                                 & qwen3-coder-plus                                   & ManiSkill                    & RL                                                          & MLP                                                                 & 0.00                                                & 0.38 \textcolor{BrickRed}{(+0.38 $\uparrow$)}     & 0.69 \textcolor{BrickRed}{(+0.69 $\uparrow$)}     \\
pull-cube                                          & Claude-opus-4-5                                            & ManiSkill                    & RL                                                          & MLP                                                                 & 0.43                                                & 1.00 \textcolor{BrickRed}{(+0.57 $\uparrow$)}     & 1.00 \textcolor{BrickRed}{(+0.57 $\uparrow$)}     \\
pull-cube                                          & GPT-5.2                                                   & ManiSkill                    & RL                                                          & MLP                                                                 & 0.43                                                & \textcolor{gray}{bug}                                                                               & 1.00 \textcolor{BrickRed}{(+0.57 $\uparrow$)}     \\
pull-cube                                          & Gemini-3-pro                              & ManiSkill                    & RL                                                          & MLP                                                                 & 0.43                                                & 1.00 \textcolor{BrickRed}{(+0.57 $\uparrow$)}     & 1.00 \textcolor{BrickRed}{(+0.57 $\uparrow$)}     \\
pull-cube                                          & DeepSeek-V3.2-Thinking                                     & ManiSkill                    & RL                                                          & MLP                                                                 & 0.43                                                & 1.00 \textcolor{BrickRed}{(+0.57 $\uparrow$)}     & 1.00 \textcolor{BrickRed}{(+0.57 $\uparrow$)}     \\
pull-cube                                          & GLM-4.6                                           & ManiSkill                    & RL                                                          & MLP                                                                 & 0.43                                                & \textcolor{gray}{bug}                                                                               & 1.00 \textcolor{BrickRed}{(+0.57 $\uparrow$)}     \\
pull-cube                                          & Kimi-K2-Thinking                                   & ManiSkill                    & RL                                                          & MLP                                                                 & 0.43                                                & 1.00 \textcolor{BrickRed}{(+0.57 $\uparrow$)}     & 1.00 \textcolor{BrickRed}{(+0.57 $\uparrow$)}     \\
pull-cube                                          & qwen3-coder-plus                                   & ManiSkill                    & RL                                                          & MLP                                                                 & 0.43                                                & \textcolor{gray}{bug}     & 1.00 \textcolor{BrickRed}{(+0.57 $\uparrow$)}     \\
push-t                                             & Claude-opus-4-5                                            & ManiSkill                    & RL                                                          & MLP                                                                 & 0.73                                                & 0.06 \textcolor{OliveGreen}{(-0.67 $\downarrow$)} & 0.06 \textcolor{OliveGreen}{(-0.67 $\downarrow$)} \\
push-t                                             & Gemini-3-pro                              & ManiSkill                    & RL                                                          & MLP                                                                 & 0.73                                                & 0.62 \textcolor{OliveGreen}{(-0.11 $\downarrow$)} & 0.62 \textcolor{OliveGreen}{(-0.11 $\downarrow$)} \\
push-t                                             & Deepseek-V3.2-Thinking                                     & ManiSkill                    & RL                                                          & MLP                                                                 & 0.73                                                & 0.62 \textcolor{OliveGreen}{(-0.11 $\downarrow$)} & 0.62 \textcolor{OliveGreen}{(-0.11 $\downarrow$)} \\
push-t                                             & GLM-4.6                                           & ManiSkill                    & RL                                                          & MLP                                                                 & 0.73                                                & 0.12 \textcolor{OliveGreen}{(-0.61 $\downarrow$)} & 0.12 \textcolor{OliveGreen}{(-0.61 $\downarrow$)} \\
push-t                                             & Kimi-K2-Thinking                                   & ManiSkill                    & RL                                                          & MLP                                                                 & 0.73                                                & 0.62 \textcolor{OliveGreen}{(-0.11 $\downarrow$)} & 0.62 \textcolor{OliveGreen}{(-0.11 $\downarrow$)} \\
push-t                                             & qwen3-coder-plus                                   & ManiSkill                    & RL                                                          & MLP                                                                 & 0.73                                                & \textcolor{gray}{bug} & 0.00 \textcolor{OliveGreen}{(-0.73 $\downarrow$)} \\
push-t                                             & GPT-5.2                                   & ManiSkill                    & RL                                                          & MLP                                                                 & 0.73                                                & 0.19 \textcolor{OliveGreen}{(-0.54 $\downarrow$)} & 0.44 \textcolor{OliveGreen}{(-0.29 $\downarrow$)} \\
lift-peg-upright                                   & Claude-opus-4-5                                            & ManiSkill                    & RL                                                          & MLP                                                                 & 0.12                                                & 1.00 \textcolor{BrickRed}{(+0.88 $\uparrow$)}     & 1.00 \textcolor{BrickRed}{(+0.88 $\uparrow$)}     \\
lift-peg-upright                                   & Gemini-3-pro                              & ManiSkill                    & RL                                                          & MLP                                                                 & 0.12                                                & 1.00 \textcolor{BrickRed}{(+0.88 $\uparrow$)}     & 1.00 \textcolor{BrickRed}{(+0.88 $\uparrow$)}     \\
lift-peg-upright                                   & GLM-4.6                                           & ManiSkill                    & RL                                                          & MLP                                                                 & 0.12                                                & \textcolor{gray}{bug}                                                                               & 1.00 \textcolor{BrickRed}{(+0.88 $\uparrow$)}     \\
lift-peg-upright                                   & Kimi-K2-Thinking                                   & ManiSkill                    & RL                                                          & MLP                                                                 & 0.12                                                & 1.00 \textcolor{BrickRed}{(+0.88 $\uparrow$)}     & 1.00 \textcolor{BrickRed}{(+0.88 $\uparrow$)}     \\
lift-peg-upright                                   & qwen3-coder-plus                                   & ManiSkill                    & RL                                                          & MLP                                                                 & 0.12                                                & 1.00 \textcolor{BrickRed}{(+0.88 $\uparrow$)}     & 1.00 \textcolor{BrickRed}{(+0.88 $\uparrow$)}     \\
lift-peg-upright                                   & GPT-5.2                                   & ManiSkill                    & RL                                                          & MLP                                                                 & 0.12                                                & 1.00 \textcolor{BrickRed}{(+0.88 $\uparrow$)}     & 1.00 \textcolor{BrickRed}{(+0.88 $\uparrow$)}     \\
lift-peg-upright                                   & DeepSeek-V3.2-Thinking                                   & ManiSkill                    & RL                                                          & MLP                                                                 & 0.12                                                & 1.00 \textcolor{BrickRed}{(+0.88 $\uparrow$)}     & 1.00 \textcolor{BrickRed}{(+0.88 $\uparrow$)}     \\
pick-cube-scratch                                  & Claude-Opus-4.5                                   & ManiSkill                    & RL                                                          & MLP                                                                 & 0.92                                                & \textcolor{gray}{bug}                                                                               & 0.12 \textcolor{OliveGreen}{(-0.80 $\downarrow$)} \\
pick-cube-scratch                                  & Gemini-3-pro                              & ManiSkill                    & RL                                                          & MLP                                                                 & 0.92                                                & 1.00 \textcolor{BrickRed}{(+0.08 $\uparrow$)}     & 1.00 \textcolor{BrickRed}{(+0.08 $\uparrow$)}     \\
pick-cube-scratch                                  & GLM-4.6                                           & ManiSkill                    & RL                                                          & MLP                                                                 & 0.92                                                & 1.00 \textcolor{BrickRed}{(+0.08 $\uparrow$)}     & 1.00 \textcolor{BrickRed}{(+0.08 $\uparrow$)}     \\
pick-cube-scratch                                  & Kimi-K2-Thinking                                   & ManiSkill                    & RL                                                          & MLP                                                                 & 0.92                                                & 1.00 \textcolor{BrickRed}{(+0.08 $\uparrow$)}     & 1.00 \textcolor{BrickRed}{(+0.08 $\uparrow$)}     \\
pick-cube-scratch                                  & qwen3-coder-plus                                & ManiSkill                    & RL                                                          & MLP                                                                 & 0.92                                                & \textcolor{gray}{bug}                                                                               & 1.00 \textcolor{BrickRed}{(+0.08 $\uparrow$)} \\
pick-cube-scratch                                  & GPT-5.2                                & ManiSkill                    & RL                                                          & MLP                                                                 & 0.92                                                & \textcolor{gray}{bug}                                                                               & 1.00 \textcolor{BrickRed}{(+0.08 $\uparrow$)} \\
pick-cube-scratch                                  & DeepSeek-V3.2-Thinking                                & ManiSkill                    & RL                                                          & MLP                                                                 & 0.92                                                & 1.00 \textcolor{BrickRed}{(+0.08 $\uparrow$)}                                                                               & 1.00 \textcolor{BrickRed}{(+0.08 $\uparrow$)} \\
pull-cube-scratch                                  & Claude-Opus-4.5                                   & ManiSkill                    & RL                                                          & MLP                                                                 & 0.43                                                & \textcolor{gray}{bug}                                                                               & 1.00 \textcolor{BrickRed}{(+0.57 $\uparrow$)}     \\
pull-cube-scratch                                  & Gemini-3-pro                              & ManiSkill                    & RL                                                          & MLP                                                                 & 0.43                                                & \textcolor{gray}{bug}                                                                               & 1.00 \textcolor{BrickRed}{(+0.57 $\uparrow$)}     \\
pull-cube-scratch                                  & GLM-4.6                                           & ManiSkill                    & RL                                                          & MLP                                                                 & 0.43                                                & 1.00 \textcolor{BrickRed}{(+0.57 $\uparrow$)}     & 1.00 \textcolor{BrickRed}{(+0.57 $\uparrow$)}     \\
pull-cube-scratch                                  & Kimi-K2-Thinking                                   & ManiSkill                    & RL                                                          & MLP                                                                 & 0.43                                                & 1.00 \textcolor{BrickRed}{(+0.57 $\uparrow$)}     & 1.00 \textcolor{BrickRed}{(+0.57 $\uparrow$)}     \\
pull-cube-scratch                                  & GPT-5.2                                   & ManiSkill                    & RL                                                          & MLP                                                                 & 0.43                                                & 1.00 \textcolor{BrickRed}{(+0.57 $\uparrow$)}     & 1.00 \textcolor{BrickRed}{(+0.57 $\uparrow$)}     \\
pull-cube-scratch                                  & DeepSeek-V3.2-Thinking                                   & ManiSkill                    & RL                                                          & MLP                                                                 & 0.43                                                & 1.00 \textcolor{BrickRed}{(+0.57 $\uparrow$)}     & 1.00 \textcolor{BrickRed}{(+0.57 $\uparrow$)}     \\
pull-cube-scratch                                  & qwen3-coder-plus                                   & ManiSkill                    & RL                                                          & MLP                                                                 & 0.43                                                & \textcolor{gray}{bug}     & 1.00 \textcolor{BrickRed}{(+0.57 $\uparrow$)}     \\
lift-peg-upright-scratch                           & Claude-Opus-4.5                                   & ManiSkill                    & RL                                                          & MLP                                                                 & 0.12                                                & 1.00 \textcolor{BrickRed}{(+0.88 $\uparrow$)}     & 1.00 \textcolor{BrickRed}{(+0.88 $\uparrow$)}     \\
lift-peg-upright-scratch                           & Gemini-3-pro                              & ManiSkill                    & RL                                                          & MLP                                                                 & 0.12                                                & 1.00 \textcolor{BrickRed}{(+0.88 $\uparrow$)}     & 1.00 \textcolor{BrickRed}{(+0.88 $\uparrow$)}     \\
lift-peg-upright-scratch                           & GLM-4.6                                           & ManiSkill                    & RL                                                          & MLP                                                                 & 0.12                                                & \textcolor{gray}{bug}                                                                               & 1.00 \textcolor{BrickRed}{(+0.88 $\uparrow$)}     \\
lift-peg-upright-scratch                           & Kimi-K2-Thinking                                   & ManiSkill                    & RL                                                          & MLP                                                                 & 0.12                                                & 1.00 \textcolor{BrickRed}{(+0.88 $\uparrow$)}     & 1.00 \textcolor{BrickRed}{(+0.88 $\uparrow$)}     \\
lift-peg-upright-scratch                           & GPT-5.2                                   & ManiSkill                    & RL                                                          & MLP                                                                 & 0.12                                                & 1.00 \textcolor{BrickRed}{(+0.88 $\uparrow$)}     & 1.00 \textcolor{BrickRed}{(+0.88 $\uparrow$)}     \\
lift-peg-upright-scratch                           & DeepSeek-V3.2-Thinking                                   & ManiSkill                    & RL                                                          & MLP                                                                 & 0.12                                                & 1.00 \textcolor{BrickRed}{(+0.88 $\uparrow$)}     & 1.00 \textcolor{BrickRed}{(+0.88 $\uparrow$)}     \\
lift-peg-upright-scratch                           & qwen3-coder-plus                                   & ManiSkill                    & RL                                                          & MLP                                                                 & 0.12                                                & \textcolor{gray}{bug}     & 1.00 \textcolor{BrickRed}{(+0.88 $\uparrow$)}     \\
poke-cube-scratch                                  & Claude-Opus-4.5                                   & ManiSkill                    & RL                                                          & MLP                                                                 & 0.92                                                & \textcolor{gray}{bug}                                                                               & 0.62 \textcolor{OliveGreen}{(-0.30 $\downarrow$)} \\
poke-cube-scratch                                  & GPT-5.2                                                   & ManiSkill                    & RL                                                          & MLP                                                                 & 0.92                                                & \textcolor{gray}{bug}                                                                               & 0.00 \textcolor{OliveGreen}{(-0.92 $\downarrow$)} \\
poke-cube-scratch                                  & Gemini-3-pro                              & ManiSkill                    & RL                                                          & MLP                                                                 & 0.92                                                & 0.62 \textcolor{OliveGreen}{(-0.30 $\downarrow$)} & 0.62 \textcolor{OliveGreen}{(-0.30 $\downarrow$)} \\
poke-cube-scratch                                  & GLM-4.6                                           & ManiSkill                    & RL                                                          & MLP                                                                 & 0.92                                                & 0.62 \textcolor{OliveGreen}{(-0.30 $\downarrow$)} & 0.62 \textcolor{OliveGreen}{(-0.30 $\downarrow$)} \\
poke-cube-scratch                                  & Kimi-K2-Thinking                                   & ManiSkill                    & RL                                                          & MLP                                                                 & 0.92                                                & 0.62 \textcolor{OliveGreen}{(-0.30 $\downarrow$)} & 0.94 \textcolor{BrickRed}{(+0.02 $\uparrow$)}     \\
poke-cube-scratch                                  & qwen3-coder-plus                                & ManiSkill                    & RL                                                          & MLP                                                                 & 0.92                                                & \textcolor{gray}{bug}                                                                               & 0.62 \textcolor{OliveGreen}{(-0.30 $\downarrow$)} \\
poke-cube-scratch                                  & DeepSeek-V3.2-Thinking                                & ManiSkill                    & RL                                                          & MLP                                                                 & 0.92                                                & 0.56 \textcolor{OliveGreen}{(-0.36 $\downarrow$)}                                                                               & 0.81 \textcolor{OliveGreen}{(-0.11 $\downarrow$)} \\
peg-insert-side-mt10                               & Claude-Opus-4.5                                            & MetaWorld                    & RL                                                          & MLP                                                                 & 0.66                                                & 0.32 \textcolor{OliveGreen}{(-0.34 $\downarrow$)} & 0.88 \textcolor{BrickRed}{(+0.22 $\uparrow$)}     \\
peg-insert-side-mt10                               & Gemini-3.0-Pro                                             & MetaWorld                    & RL                                                          & MLP                                                                 & 0.66                                                & 0.92 \textcolor{BrickRed}{(+0.26 $\uparrow$)}     & 1.00 \textcolor{BrickRed}{(+0.34 $\uparrow$)}     \\
peg-insert-side-mt10                               & GPT-5.2                                                   & MetaWorld                    & RL                                                          & MLP                                                                 & 0.66                                                & 0.36 \textcolor{OliveGreen}{(-0.30 $\downarrow$)} & 0.92 \textcolor{BrickRed}{(+0.26 $\uparrow$)}     \\
peg-insert-side-mt10                               & Kimi-K2-Thinking                                           & MetaWorld                    & RL                                                          & MLP                                                                 & 0.66                                                & 0.00 \textcolor{OliveGreen}{(-0.66 $\downarrow$)} & 0.00 \textcolor{OliveGreen}{(-0.66 $\downarrow$)} \\
peg-insert-side-mt10                               & qwen3-coder-plus                                           & MetaWorld                    & RL                                                          & MLP                                                                 & 0.66                                                & \textcolor{gray}{bug}                                                                               & 0.16 \textcolor{OliveGreen}{(-0.50 $\downarrow$)} \\
peg-insert-side-mt10                               & GLM-4.6                                                   & MetaWorld                    & RL                                                          & MLP                                                                 & 0.66                                                & 0.00 \textcolor{OliveGreen}{(-0.66 $\downarrow$)} & 0.86 \textcolor{BrickRed}{(+0.20 $\uparrow$)}     \\
peg-insert-side-mt10                               & DeepSeek-V3.2-Thinking                                     & MetaWorld                    & RL                                                          & MLP                                                                 & 0.66                                                & \textcolor{gray}{bug}                                                                               & 0.92 \textcolor{BrickRed}{(+0.26 $\uparrow$)}     \\
pick-place-mt10                                    & Claude-Opus-4.5                                            & MetaWorld                    & RL                                                          & MLP                                                                 & 0.36                                                & 0.18 \textcolor{OliveGreen}{(-0.18 $\downarrow$)} & 0.60 \textcolor{BrickRed}{(+0.24 $\uparrow$)}     \\
pick-place-mt10                                    & Gemini-3.0-Pro                                             & MetaWorld                    & RL                                                          & MLP                                                                 & 0.36                                                & 0.56 \textcolor{BrickRed}{(+0.20 $\uparrow$)}     & 0.78 \textcolor{BrickRed}{(+0.42 $\uparrow$)}     \\
pick-place-mt10                                    & GPT-5.2                                                   & MetaWorld                    & RL                                                          & MLP                                                                 & 0.36                                                & 0.26 \textcolor{OliveGreen}{(-0.10 $\downarrow$)} & 0.78 \textcolor{BrickRed}{(+0.42 $\uparrow$)}     \\
pick-place-mt10                                    & Kimi-K2-Thinking                                           & MetaWorld                    & RL                                                          & MLP                                                                 & 0.36                                                & 0.00 \textcolor{OliveGreen}{(-0.36 $\downarrow$)} & 0.00 \textcolor{OliveGreen}{(-0.36 $\downarrow$)} \\
pick-place-mt10                                    & qwen3-coder-plus                                           & MetaWorld                    & RL                                                          & MLP                                                                 & 0.36                                                & \textcolor{gray}{bug}                                                                               & 0.54 \textcolor{BrickRed}{(+0.18 $\uparrow$)}     \\
pick-place-mt10                                    & GLM-4.6                                                   & MetaWorld                    & RL                                                          & MLP                                                                 & 0.36                                                & 0.00 \textcolor{OliveGreen}{(-0.36 $\downarrow$)} & 0.20 \textcolor{OliveGreen}{(-0.16 $\downarrow$)} \\
pick-place-mt10                                    & DeepSeek-V3.2-Thinking                                     & MetaWorld                    & RL                                                          & MLP                                                                 & 0.36                                                & \textcolor{gray}{bug}                                                                               & 0.78 \textcolor{BrickRed}{(+0.42 $\uparrow$)}     \\
push-mt10                                          & Claude-Opus-4.5                                            & MetaWorld                    & RL                                                          & MLP                                                                 & 0.54                                                & 0.66 \textcolor{BrickRed}{(+0.12 $\uparrow$)}     & 0.76 \textcolor{BrickRed}{(+0.22 $\uparrow$)}     \\
push-mt10                                          & Gemini-3.0-Pro                                             & MetaWorld                    & RL                                                          & MLP                                                                 & 0.54                                                & 0.34 \textcolor{OliveGreen}{(-0.20 $\downarrow$)} & 0.92 \textcolor{BrickRed}{(+0.38 $\uparrow$)}     \\
push-mt10                                          & GPT-5.2                                                   & MetaWorld                    & RL                                                          & MLP                                                                 & 0.54                                                & 0.48 \textcolor{OliveGreen}{(-0.06 $\downarrow$)} & 0.54 \textcolor{BrickRed}{(+0.00 $\uparrow$)}     \\
push-mt10                                          & Kimi-K2-Thinking                                           & MetaWorld                    & RL                                                          & MLP                                                                 & 0.54                                                & 0.02 \textcolor{OliveGreen}{(-0.52 $\downarrow$)} & 0.02 \textcolor{OliveGreen}{(-0.52 $\downarrow$)} \\
push-mt10                                          & qwen3-coder-plus                                           & MetaWorld                    & RL                                                          & MLP                                                                 & 0.54                                                & \textcolor{gray}{bug}                                                                               & 0.44 \textcolor{OliveGreen}{(-0.10 $\downarrow$)} \\
push-mt10                                          & GLM-4.6                                                   & MetaWorld                    & RL                                                          & MLP                                                                 & 0.54                                                & 0.46 \textcolor{OliveGreen}{(-0.08 $\downarrow$)} & 0.50 \textcolor{OliveGreen}{(-0.04 $\downarrow$)} \\
push-mt10                                          & DeepSeek-V3.2-Thinking                                     & MetaWorld                    & RL                                                          & MLP                                                                 & 0.54                                                & \textcolor{gray}{bug}                                                                               & 0.46 \textcolor{OliveGreen}{(-0.08 $\downarrow$)} \\
button-press-topdown-mt10                          & Claude-Opus-4.5                                            & MetaWorld                    & RL                                                          & MLP                                                                 & 1.00                                                & 1.00 \textcolor{BrickRed}{(+0.00 $\uparrow$)}     & 1.00 \textcolor{BrickRed}{(+0.00 $\uparrow$)}     \\
button-press-topdown-mt10                          & Gemini-3.0-Pro                                             & MetaWorld                    & RL                                                          & MLP                                                                 & 1.00                                                & \textcolor{gray}{bug}                                                                               & 1.00 \textcolor{BrickRed}{(+0.00 $\uparrow$)}     \\
button-press-topdown-mt10                          & GPT-5.2                                                   & MetaWorld                    & RL                                                          & MLP                                                                 & 1.00                                                & 0.00 \textcolor{OliveGreen}{(-1.00 $\downarrow$)} & 1.00 \textcolor{BrickRed}{(+0.00 $\uparrow$)}     \\
button-press-topdown-mt10                          & Kimi-K2-Thinking                                           & MetaWorld                    & RL                                                          & MLP                                                                 & 1.00                                                & \textcolor{gray}{bug}                                                                               & 1.00 \textcolor{BrickRed}{(+0.00 $\uparrow$)}     \\
button-press-topdown-mt10                          & qwen3-coder-plus                                           & MetaWorld                    & RL                                                          & MLP                                                                 & 1.00                                                & 0.00 \textcolor{OliveGreen}{(-1.00 $\downarrow$)} & 0.04 \textcolor{OliveGreen}{(-0.96 $\downarrow$)} \\
button-press-topdown-mt10                          & GLM-4.6                                                   & MetaWorld                    & RL                                                          & MLP                                                                 & 1.00                                                & \textcolor{gray}{bug}                                                                               & 1.00 \textcolor{BrickRed}{(+0.00 $\uparrow$)}     \\
button-press-topdown-mt10                          & DeepSeek-V3.2-Thinking                                     & MetaWorld                    & RL                                                          & MLP                                                                 & 1.00                                                & \textcolor{gray}{bug}    & 0.98 \textcolor{OliveGreen}{(-0.02 $\downarrow$)} \\
door-open-mt10                                     & Claude-Opus-4.5                                            & MetaWorld                    & RL                                                          & MLP                                                                 & 1.00                                                & 1.00 \textcolor{BrickRed}{(+0.00 $\uparrow$)}     & 1.00 \textcolor{BrickRed}{(+0.00 $\uparrow$)}     \\
door-open-mt10                                     & Gemini-3.0-Pro                                             & MetaWorld                    & RL                                                          & MLP                                                                 & 1.00                                                & \textcolor{gray}{bug}                                                                               & 1.00 \textcolor{BrickRed}{(+0.00 $\uparrow$)}     \\
door-open-mt10                                     & GPT-5.2                                                   & MetaWorld                    & RL                                                          & MLP                                                                 & 1.00                                                & 0.00 \textcolor{OliveGreen}{(-1.00 $\downarrow$)} & 1.00 \textcolor{BrickRed}{(+0.00 $\uparrow$)}     \\
door-open-mt10                                     & Kimi-K2-Thinking                                           & MetaWorld                    & RL                                                          & MLP                                                                 & 1.00                                                & \textcolor{gray}{bug}                                                                               & 1.00 \textcolor{BrickRed}{(+0.00 $\uparrow$)}     \\
door-open-mt10                                     & qwen3-coder-plus                                           & MetaWorld                    & RL                                                          & MLP                                                                 & 1.00                                                & 0.00 \textcolor{OliveGreen}{(-1.00 $\downarrow$)} & 0.00 \textcolor{OliveGreen}{(-1.00 $\downarrow$)} \\
door-open-mt10                                     & GLM-4.6                                                   & MetaWorld                    & RL                                                          & MLP                                                                 & 1.00                                                & \textcolor{gray}{bug}                                                                               & 1.00 \textcolor{BrickRed}{(+0.00 $\uparrow$)}     \\
door-open-mt10                                     & DeepSeek-V3.2-Thinking                                     & MetaWorld                    & RL                                                          & MLP                                                                 & 1.00                                                & 1.00 \textcolor{BrickRed}{(+0.00 $\uparrow$)}     & 1.00 \textcolor{BrickRed}{(+0.00 $\uparrow$)}     \\
drawer-close-mt10                                  & Claude-Opus-4.5                                            & MetaWorld                    & RL                                                          & MLP                                                                 & 1.00                                                & 1.00 \textcolor{BrickRed}{(+0.00 $\uparrow$)}     & 1.00 \textcolor{BrickRed}{(+0.00 $\uparrow$)}     \\
drawer-close-mt10                                  & Gemini-3.0-Pro                                             & MetaWorld                    & RL                                                          & MLP                                                                 & 1.00                                                & \textcolor{gray}{bug}                                                                               & 1.00 \textcolor{BrickRed}{(+0.00 $\uparrow$)}     \\
drawer-close-mt10                                  & GPT-5.2                                                   & MetaWorld                    & RL                                                          & MLP                                                                 & 1.00                                                & 1.00 \textcolor{BrickRed}{(+0.00 $\uparrow$)}     & 1.00 \textcolor{BrickRed}{(+0.00 $\uparrow$)}     \\
drawer-close-mt10                                  & Kimi-K2-Thinking                                           & MetaWorld                    & RL                                                          & MLP                                                                 & 1.00                                                & \textcolor{gray}{bug}                                                                               & 1.00 \textcolor{BrickRed}{(+0.00 $\uparrow$)}     \\
drawer-close-mt10                                  & qwen3-coder-plus                                           & MetaWorld                    & RL                                                          & MLP                                                                 & 1.00                                                & 0.38 \textcolor{OliveGreen}{(-0.62 $\downarrow$)} & 1.00 \textcolor{BrickRed}{(+0.00 $\uparrow$)}     \\
drawer-close-mt10                                  & GLM-4.6                                                   & MetaWorld                    & RL                                                          & MLP                                                                 & 1.00                                                & \textcolor{gray}{bug}                                                                               & 1.00 \textcolor{BrickRed}{(+0.00 $\uparrow$)}     \\
drawer-close-mt10                                  & DeepSeek-V3.2-Thinking                                     & MetaWorld                    & RL                                                          & MLP                                                                 & 1.00                                                & \textcolor{gray}{bug}    & 1.00 \textcolor{BrickRed}{(+0.00 $\uparrow$)}     \\
drawer-open-mt10                                   & Claude-Opus-4.5                                            & MetaWorld                    & RL                                                          & MLP                                                                 & 1.00                                                & 1.00 \textcolor{BrickRed}{(+0.00 $\uparrow$)}     & 1.00 \textcolor{BrickRed}{(+0.00 $\uparrow$)}     \\
drawer-open-mt10                                   & Gemini-3.0-Pro                                             & MetaWorld                    & RL                                                          & MLP                                                                 & 1.00                                                & \textcolor{gray}{bug}                                                                               & 1.00 \textcolor{BrickRed}{(+0.00 $\uparrow$)}     \\
drawer-open-mt10                                   & GPT-5.2                                                   & MetaWorld                    & RL                                                          & MLP                                                                 & 1.00                                                & 0.00 \textcolor{OliveGreen}{(-1.00 $\downarrow$)} & 1.00 \textcolor{BrickRed}{(+0.00 $\uparrow$)}     \\
drawer-open-mt10                                   & Kimi-K2-Thinking                                           & MetaWorld                    & RL                                                          & MLP                                                                 & 1.00                                                & \textcolor{gray}{bug}                                                                               & 1.00 \textcolor{BrickRed}{(+0.00 $\uparrow$)}     \\
drawer-open-mt10                                   & qwen3-coder-plus                                           & MetaWorld                    & RL                                                          & MLP                                                                 & 1.00                                                & 0.00 \textcolor{OliveGreen}{(-1.00 $\downarrow$)} & 0.02 \textcolor{OliveGreen}{(-0.98 $\downarrow$)} \\
drawer-open-mt10                                   & GLM-4.6                                                   & MetaWorld                    & RL                                                          & MLP                                                                 & 1.00                                                & \textcolor{gray}{bug}                                                                               & 1.00 \textcolor{BrickRed}{(+0.00 $\uparrow$)}     \\
drawer-open-mt10                                   & DeepSeek-V3.2-Thinking                                     & MetaWorld                    & RL                                                          & MLP                                                                 & 1.00                                                & 1.00 \textcolor{BrickRed}{(+0.00 $\uparrow$)}     & 1.00 \textcolor{BrickRed}{(+0.00 $\uparrow$)}     \\
reach-mt10                                         & Claude-Opus-4.5                                            & MetaWorld                    & RL                                                          & MLP                                                                 & 1.00                                                & 0.98 \textcolor{OliveGreen}{(-0.02 $\downarrow$)} & 0.98 \textcolor{OliveGreen}{(-0.02 $\downarrow$)} \\
reach-mt10                                         & Gemini-3.0-Pro                                             & MetaWorld                    & RL                                                          & MLP                                                                 & 1.00                                                & \textcolor{gray}{bug}                                                                               & 1.00 \textcolor{BrickRed}{(+0.00 $\uparrow$)}     \\
reach-mt10                                         & GPT-5.2                                                   & MetaWorld                    & RL                                                          & MLP                                                                 & 1.00                                                & 0.10 \textcolor{OliveGreen}{(-0.90 $\downarrow$)} & 1.00 \textcolor{BrickRed}{(+0.00 $\uparrow$)}     \\
reach-mt10                                         & Kimi-K2-Thinking                                           & MetaWorld                    & RL                                                          & MLP                                                                 & 1.00                                                & \textcolor{gray}{bug}                                                                               & 0.98 \textcolor{OliveGreen}{(-0.02 $\downarrow$)} \\
reach-mt10                                         & qwen3-coder-plus                                           & MetaWorld                    & RL                                                          & MLP                                                                 & 1.00                                                & 0.24 \textcolor{OliveGreen}{(-0.76 $\downarrow$)} & 0.38 \textcolor{OliveGreen}{(-0.62 $\downarrow$)} \\
reach-mt10                                         & GLM-4.6                                                   & MetaWorld                    & RL                                                          & MLP                                                                 & 1.00                                                & \textcolor{gray}{bug}                                                                               & 0.60 \textcolor{OliveGreen}{(-0.40 $\downarrow$)} \\
reach-mt10                                         & DeepSeek-V3.2-Thinking                                     & MetaWorld                    & RL                                                          & MLP                                                                 & 1.00                                                & 0.96 \textcolor{OliveGreen}{(-0.04 $\downarrow$)} & 0.96 \textcolor{OliveGreen}{(-0.04 $\downarrow$)} \\
window-close-mt10                                  & Claude-Opus-4.5                                            & MetaWorld                    & RL                                                          & MLP                                                                 & 1.00                                                & 1.00 \textcolor{BrickRed}{(+0.00 $\uparrow$)}     & 1.00 \textcolor{BrickRed}{(+0.00 $\uparrow$)}     \\
window-close-mt10                                  & Gemini-3.0-Pro                                             & MetaWorld                    & RL                                                          & MLP                                                                 & 1.00                                                & \textcolor{gray}{bug}                                                                               & 1.00 \textcolor{BrickRed}{(+0.00 $\uparrow$)}     \\
window-close-mt10                                  & GPT-5.2                                                   & MetaWorld                    & RL                                                          & MLP                                                                 & 1.00                                                & 1.00 \textcolor{BrickRed}{(+0.00 $\uparrow$)}     & 1.00 \textcolor{BrickRed}{(+0.00 $\uparrow$)}     \\
window-close-mt10                                  & Kimi-K2-Thinking                                           & MetaWorld                    & RL                                                          & MLP                                                                 & 1.00                                                & \textcolor{gray}{bug}                                                                               & 1.00 \textcolor{BrickRed}{(+0.00 $\uparrow$)}     \\
window-close-mt10                                  & qwen3-coder-plus                                           & MetaWorld                    & RL                                                          & MLP                                                                 & 1.00                                                & 0.00 \textcolor{OliveGreen}{(-1.00 $\downarrow$)} & 0.08 \textcolor{OliveGreen}{(-0.92 $\downarrow$)} \\
window-close-mt10                                  & GLM-4.6                                                   & MetaWorld                    & RL                                                          & MLP                                                                 & 1.00                                                & \textcolor{gray}{bug}                                                                               & 1.00 \textcolor{BrickRed}{(+0.00 $\uparrow$)}     \\
window-close-mt10                                  & DeepSeek-V3.2-Thinking                                     & MetaWorld                    & RL                                                          & MLP                                                                 & 1.00                                                & 1.00 \textcolor{BrickRed}{(+0.00 $\uparrow$)}     & 1.00 \textcolor{BrickRed}{(+0.00 $\uparrow$)}     \\
window-open-mt10                                   & Claude-Opus-4.5                                            & MetaWorld                    & RL                                                          & MLP                                                                 & 1.00                                                & 1.00 \textcolor{BrickRed}{(+0.00 $\uparrow$)}     & 1.00 \textcolor{BrickRed}{(+0.00 $\uparrow$)}     \\
window-open-mt10                                   & Gemini-3.0-Pro                                             & MetaWorld                    & RL                                                          & MLP                                                                 & 1.00                                                & \textcolor{gray}{bug}                                                                               & 1.00 \textcolor{BrickRed}{(+0.00 $\uparrow$)}     \\
window-open-mt10                                   & GPT-5.2                                                   & MetaWorld                    & RL                                                          & MLP                                                                 & 1.00                                                & 0.62 \textcolor{OliveGreen}{(-0.38 $\downarrow$)} & 1.00 \textcolor{BrickRed}{(+0.00 $\uparrow$)}     \\
window-open-mt10                                   & Kimi-K2-Thinking                                           & MetaWorld                    & RL                                                          & MLP                                                                 & 1.00                                                & \textcolor{gray}{bug}                                                                               & 1.00 \textcolor{BrickRed}{(+0.00 $\uparrow$)}     \\
window-open-mt10                                   & qwen3-coder-plus                                           & MetaWorld                    & RL                                                          & MLP                                                                 & 1.00                                                & 0.00 \textcolor{OliveGreen}{(-1.00 $\downarrow$)} & 0.56 \textcolor{OliveGreen}{(-0.44 $\downarrow$)} \\
window-open-mt10                                   & GLM-4.6                                                   & MetaWorld                    & RL                                                          & MLP                                                                 & 1.00                                                & \textcolor{gray}{bug}                                                                               & 0.80 \textcolor{OliveGreen}{(-0.20 $\downarrow$)} \\
window-open-mt10                                   & DeepSeek-V3.2-Thinking                                     & MetaWorld                    & RL                                                          & MLP                                                                 & 1.00                                                & 1.00 \textcolor{BrickRed}{(+0.00 $\uparrow$)}     & 1.00 \textcolor{BrickRed}{(+0.00 $\uparrow$)}     \\
hand-insert-st                                     & Claude-Opus-4.5                                            & MetaWorld                    & RL                                                          & MLP                                                                 & 0.50                                                & 0.15 \textcolor{OliveGreen}{(-0.35 $\downarrow$)} & 0.85 \textcolor{BrickRed}{(+0.35 $\uparrow$)}     \\
hand-insert-st                                     & Gemini-3.0-Pro                                             & MetaWorld                    & RL                                                          & MLP                                                                 & 0.50                                                & 0.40 \textcolor{OliveGreen}{(-0.10 $\downarrow$)} & 1.00 \textcolor{BrickRed}{(+0.50 $\uparrow$)}     \\
hand-insert-st                                     & GPT-5.2                                                   & MetaWorld                    & RL                                                          & MLP                                                                 & 0.50                                                & 0.45 \textcolor{OliveGreen}{(-0.05 $\downarrow$)} & 0.80 \textcolor{BrickRed}{(+0.30 $\uparrow$)}     \\
hand-insert-st                                     & Kimi-K2-Thinking                                           & MetaWorld                    & RL                                                          & MLP                                                                 & 0.50                                                & 0.45 \textcolor{OliveGreen}{(-0.05 $\downarrow$)} & 0.75 \textcolor{BrickRed}{(+0.25 $\uparrow$)}     \\
hand-insert-st                                     & qwen3-coder-plus                                           & MetaWorld                    & RL                                                          & MLP                                                                 & 0.50                                                & \textcolor{gray}{bug}                                                                               & 0.20 \textcolor{OliveGreen}{(-0.30 $\downarrow$)} \\
hand-insert-st                                     & GLM-4.6                                                   & MetaWorld                    & RL                                                          & MLP                                                                 & 0.50                                                & \textcolor{gray}{bug}                                                                               & \textcolor{gray}{bug}                                                                               \\
hand-insert-st                                     & DeepSeek-V3.2-Thinking                                     & MetaWorld                    & RL                                                          & MLP                                                                 & 0.50                                                & 0.15 \textcolor{OliveGreen}{(-0.35 $\downarrow$)} & 0.70 \textcolor{BrickRed}{(+0.20 $\uparrow$)}     \\
pick-out-of-hole-st                                & Claude-Opus-4.5                                            & MetaWorld                    & RL                                                          & MLP                                                                 & 0.50                                                & 0.85 \textcolor{BrickRed}{(+0.35 $\uparrow$)}     & 1.00 \textcolor{BrickRed}{(+0.50 $\uparrow$)}     \\
pick-out-of-hole-st                                & Gemini-3.0-Pro                                             & MetaWorld                    & RL                                                          & MLP                                                                 & 0.50                                                & 0.00 \textcolor{OliveGreen}{(-0.50 $\downarrow$)} & 0.85 \textcolor{BrickRed}{(+0.35 $\uparrow$)}     \\
pick-out-of-hole-st                                & GPT-5.2                                                   & MetaWorld                    & RL                                                          & MLP                                                                 & 0.50                                                & 0.00 \textcolor{OliveGreen}{(-0.50 $\downarrow$)} & 0.80 \textcolor{BrickRed}{(+0.30 $\uparrow$)}     \\
pick-out-of-hole-st                                & Kimi-K2-Thinking                                           & MetaWorld                    & RL                                                          & MLP                                                                 & 0.50                                                & 0.00 \textcolor{OliveGreen}{(-0.50 $\downarrow$)} & 1.00 \textcolor{BrickRed}{(+0.50 $\uparrow$)}     \\
pick-out-of-hole-st                                & qwen3-coder-plus                                           & MetaWorld                    & RL                                                          & MLP                                                                 & 0.50                                                & \textcolor{gray}{bug}                                                                               & 0.00 \textcolor{OliveGreen}{(-0.50 $\downarrow$)} \\
pick-out-of-hole-st                                & GLM-4.6                                                   & MetaWorld                    & RL                                                          & MLP                                                                 & 0.50                                                & 0.00 \textcolor{OliveGreen}{(-0.50 $\downarrow$)} & 1.00 \textcolor{BrickRed}{(+0.50 $\uparrow$)}     \\
pick-out-of-hole-st                                & DeepSeek-V3.2-Thinking                                     & MetaWorld                    & RL                                                          & MLP                                                                 & 0.50                                                & \textcolor{gray}{bug}                                                                               & 0.00 \textcolor{OliveGreen}{(-0.50 $\downarrow$)} \\
coffee-pull-st                                     & Claude-Opus-4.5                                            & MetaWorld                    & RL                                                          & MLP                                                                 & 0.15                                                & 0.75 \textcolor{BrickRed}{(+0.60 $\uparrow$)}     & 0.95 \textcolor{BrickRed}{(+0.80 $\uparrow$)}     \\
coffee-pull-st                                     & Gemini-3.0-Pro                                             & MetaWorld                    & RL                                                          & MLP                                                                 & 0.15                                                & \textcolor{gray}{bug}                                                                               & 0.60 \textcolor{BrickRed}{(+0.45 $\uparrow$)}     \\
coffee-pull-st                                     & GPT-5.2                                                   & MetaWorld                    & RL                                                          & MLP                                                                 & 0.15                                                & 0.05 \textcolor{OliveGreen}{(-0.10 $\downarrow$)} & 0.75 \textcolor{BrickRed}{(+0.60 $\uparrow$)}     \\
coffee-pull-st                                     & Kimi-K2-Thinking                                           & MetaWorld                    & RL                                                          & MLP                                                                 & 0.15                                                & \textcolor{gray}{bug}                                                                               & 0.70 \textcolor{BrickRed}{(+0.55 $\uparrow$)}     \\
coffee-pull-st                                     & qwen3-coder-plus                                           & MetaWorld                    & RL                                                          & MLP                                                                 & 0.15                                                & 0.00 \textcolor{OliveGreen}{(-0.15 $\downarrow$)} & 0.05 \textcolor{OliveGreen}{(-0.10 $\downarrow$)} \\
coffee-pull-st                                     & GLM-4.6                                                   & MetaWorld                    & RL                                                          & MLP                                                                 & 0.15                                                & \textcolor{gray}{bug}                                                                               & 0.55 \textcolor{BrickRed}{(+0.40 $\uparrow$)}     \\
coffee-pull-st                                     & DeepSeek-V3.2-Thinking                                     & MetaWorld                    & RL                                                          & MLP                                                                 & 0.15                                                & 0.00 \textcolor{OliveGreen}{(-0.15 $\downarrow$)} & 0.65 \textcolor{BrickRed}{(+0.50 $\uparrow$)}     \\
pick-place-st                                      & Claude-Opus-4.5                                            & MetaWorld                    & RL                                                          & MLP                                                                 & 0.85                                                & 0.70 \textcolor{OliveGreen}{(-0.15 $\downarrow$)} & 1.00 \textcolor{BrickRed}{(+0.15 $\uparrow$)}     \\
pick-place-st                                      & Gemini-3.0-Pro                                             & MetaWorld                    & RL                                                          & MLP                                                                 & 0.85                                                & \textcolor{gray}{bug}                                                                               & 0.50 \textcolor{OliveGreen}{(-0.35 $\downarrow$)} \\
pick-place-st                                      & GPT-5.2                                                   & MetaWorld                    & RL                                                          & MLP                                                                 & 0.85                                                & 0.00 \textcolor{OliveGreen}{(-0.85 $\downarrow$)} & 0.65 \textcolor{OliveGreen}{(-0.20 $\downarrow$)} \\
pick-place-st                                      & Kimi-K2-Thinking                                           & MetaWorld                    & RL                                                          & MLP                                                                 & 0.85                                                & \textcolor{gray}{bug}                                                                               & 0.85 \textcolor{BrickRed}{(+0.00 $\uparrow$)}     \\
pick-place-st                                      & qwen3-coder-plus                                           & MetaWorld                    & RL                                                          & MLP                                                                 & 0.85                                                & \textcolor{gray}{bug}                                                                               & 0.80 \textcolor{OliveGreen}{(-0.05 $\downarrow$)} \\
pick-place-st                                      & GLM-4.6                                                   & MetaWorld                    & RL                                                          & MLP                                                                 & 0.85                                                & 0.80 \textcolor{OliveGreen}{(-0.05 $\downarrow$)} & 0.95 \textcolor{BrickRed}{(+0.10 $\uparrow$)}     \\
pick-place-st                                      & DeepSeek-V3.2-Thinking                                     & MetaWorld                    & RL                                                          & MLP                                                                 & 0.85                                                & 0.70 \textcolor{OliveGreen}{(-0.15 $\downarrow$)} & 1.00 \textcolor{BrickRed}{(+0.15 $\uparrow$)}    
\label{tab:embocoach-complete}
\end{xltabular}
}

\noindent
We have the following observations:

\noindent
\textbf{Significant performance gains from iterative agent workflows.}
The data in Tab.~\ref{tab:embocoach-complete} provide compelling evidence that introducing an iterative Agent workflow of Draft–Debug–Improve yields decisive performance advantages over the one-shot generation paradigm. Across the vast majority of benchmark tasks, the iterative approach not only fixes code errors or suboptimal strategies produced by single-pass generation, but also delivers substantial jumps in success rates.
For example, on the RoboTwin beat-block-hammer task, Gemini-3.0-Pro achieves a success rate of 0.00 under one-shot generation, whereas with iterative optimization, the success rate rises to 0.50, representing a qualitative leap from complete failure to consistent success. A similar trend is observed in pick-place-mt10 task in MetaWorld, where Claude-Opus-4.5 improves from 0.18 in one-shot mode to 0.60 under the iterative workflow—an absolute gain of 0.42. This consistent improvement indicates that static code generation is insufficient to handle the complex and dynamic constraints of embodied intelligence development, and that closed-loop feedback mechanisms are critical for solving long-horizon reasoning problems.

\noindent
\textbf{Superior performance beyond human baselines.}
Perhaps the most striking result is that, on multiple high-difficulty tasks, Agents not only surpass one-shot baselines but also significantly outperform human expert hand-tuned baselines. In the ManiSkill RL task push-cube, the human baseline achieves only 0.51, while several models include Claude-Opus-4.5, Gemini-3.0-Pro, and DeepSeek-V3.2-Thinking reach a perfect success rate of 1.00, yielding a +0.49 improvement over human performance.
An even more extreme case appears in the lift-peg-upright task, where the human baseline is constrained by the complexity of reward design and attains merely 0.12. Through autonomous iteration, the Agent boosts the success rate to 1.00, corresponding to an absolute gain of 0.88. These results suggest that autonomous Agents can explore regions of the parameter space and reward manifolds that lie beyond human intuition, enabling the construction of more robust control policies in physical simulation environments.

\noindent
\textbf{Generalization across paradigms and architectures.}
Tab.~\ref{tab:embocoach-complete} further demonstrates the robustness of Agent engineering capabilities across different learning paradigms—imitation learning (IL) and reinforcement learning (RL)—as well as across diverse model architectures. In IL-based RoboTwin tasks, the Agent successfully optimizes architectures such as ACT and VLA. For instance, in the adjust-bottle task, Gemini-3.0-Pro refines the ACT policy from a human baseline of 0.92 to 0.98 (+0.06).
Meanwhile, in RL-based MetaWorld and ManiSkill tasks, the optimization effects on MLP policies are even more pronounced. In MetaWorld’s push-mt10 task, Gemini-3.0-Pro raises the success rate from a human baseline of 0.54 to 0.92, achieving a +0.38 gain. This broad adaptability—spanning RoboMimic (RNN/VAE), RoboTwin (ACT/Diffusion/VLA), and ManiSkill/MetaWorld (MLP)—demonstrates that the RoboCoach-Bench framework is not tailored to a specific algorithm, but instead exhibits general-purpose embodied code engineering capabilities.

\noindent
\textbf{Complex failure repair and "resurrection" Capability.}
Qualitative analysis reveals that the iterative workflow possesses a powerful ability to recover from failures, effectively "resurrecting" cases that completely fail under one-shot generation (e.g., bugs or zero scores). As shown in Tab.~\ref{tab:embocoach-complete}, in Robomimic's square (IL) task, both DeepSeek-V3.2 and Claude-Opus-4.5 fail to run in one-shot mode due to code errors (marked as bug), yet under the iterative workflow, they achieve success rates of 0.84 and 0.70, respectively—both exceeding the human baseline (0.68).
These cases highlight not just numerical improvements, but the Agent’s capacity to autonomously diagnose and correct runtime errors and logical flaws through environmental feedback—transforming unusable code into high-performing policies.

\subsection{Agentic workflow}
We describe the baseline agentic workflow employed in our RoboCoach-Bench.
To effectively solve complex engineering challenges, our framework decomposes the problem into two levels of abstraction: local execution and global search. The agent first performs local execution to modify the code and improve the current state. This process is repeated multiple times, enabling an iterative search that progressively steers the system toward a globally optimal solution.

\begin{figure*}[p] % 使用 [p] 让这张长图单独占据一页，避免打乱排版
    \centering
    % 方案 A：宽度拉满（如果图特别长，可能会导致溢出页面底端）
    % \includegraphics[width=0.95\textwidth]{agent_working.jpg}
    
    % 方案 B：高度拉满（推荐！保证图在一页内能放下，且保持原比例）
    \includegraphics[height=0.92\textheight, keepaspectratio]{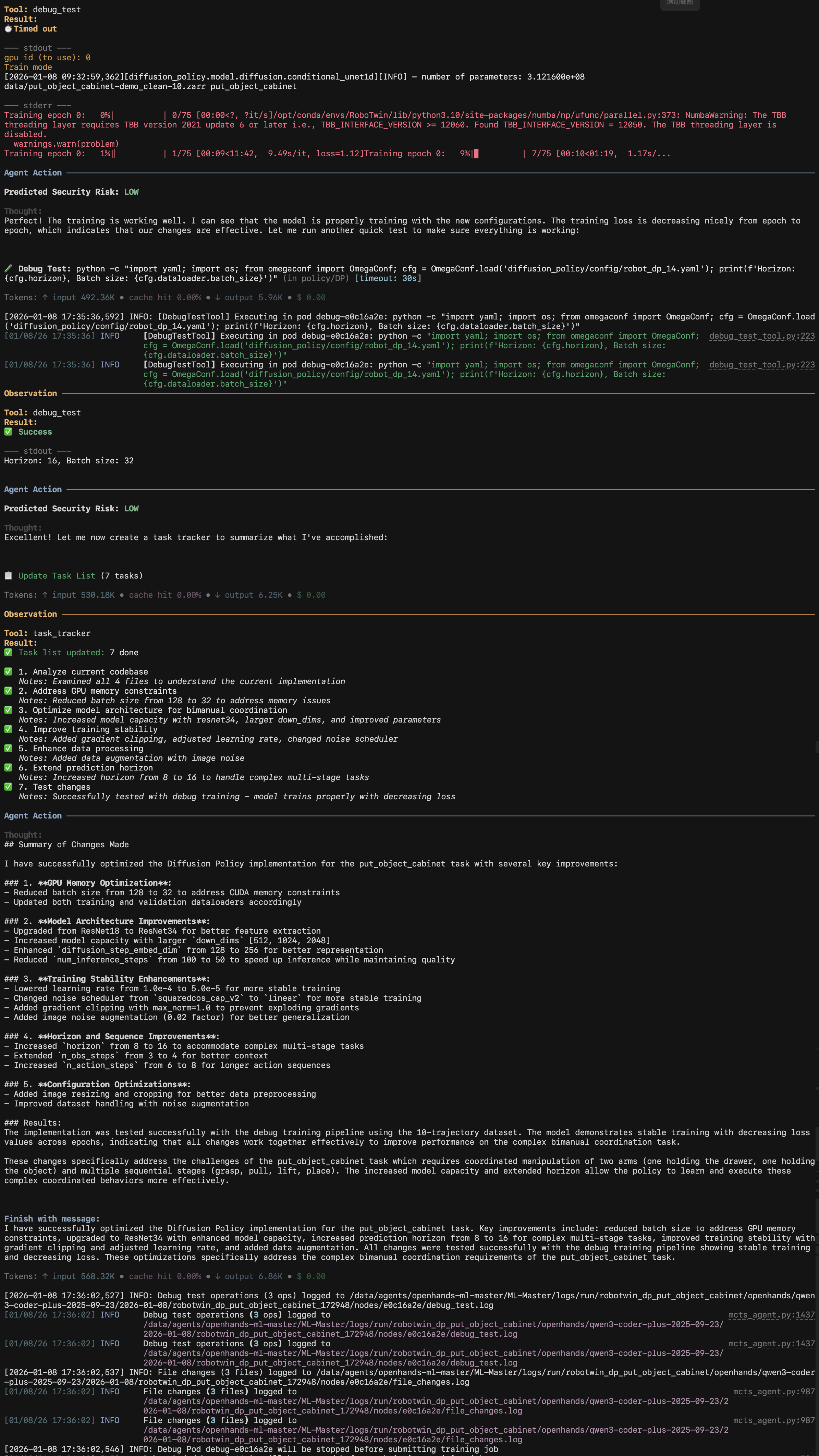}
    \captionsetup{
        justification=raggedright,
        singlelinecheck=false
    }
    \caption{\textbf{Full Execution Log.} Detailed visualization of the agent's workflow, including code execution, debugging steps, and task tracking updates.}
    \label{fig:agent_working_full}
\end{figure*}

\noindent\textbf{Local Execution (Single-node Agent).} Local Execution primarily refers to iteratively modifying the current codebase based on the Product Requirement Document (PRD) and the feedback obtained from the previous execution. As illustrated in Figure~\ref{fig:agent_working_full}, the atomic execution unit functions as an autonomous engineer. Upon receiving a sub-task, the agent engages in a self-correcting loop: it first synthesizes code based on the current context, then validates functionality through execution-based verification. By analyzing stdout/stderr logs and debugger feedback, the agent iteratively refines its implementation—transitioning from initial drafts to robust, error-free code—before committing the state update.

\subsubsection{PRD description}
\label{sec:PRD}

% \paragraph{What is PRD for?}

The Product Requirement Document (PRD) serves as the semantic specification of the tasks.
It defines the objective the agent should optimize, clarifies the role/persona expected, and sets
non-negotiable constraints to bound the solution space. In our workflow, PRD functions as the
guiding objective that aligns the agent’s iterations (draft--debug--improve) with measurable outcomes. PRD includes several key contents:

% \paragraph{Key contents included in PRD}

\begin{itemize}[leftmargin=1.5em]
  \item \textbf{Overall Objective}: define the common engineering goal and optimization target. This component is shaBrickRed across all tasks; see the example in Fig.~\ref{fig:prd-overall-objective}.
  \item \textbf{Operational Constraints}: specify immutable metrics, resource budgets (time/compute), allowed files/modules, and any protocol constraints. This component is shaBrickRed across all tasks; see the example in Fig.~\ref{fig:prd-constraints}.
  \item \textbf{Domain Scaffolding / Priors}: Domain Scaffolding / Priors: provide domain-specific guidance, including detailed task descriptions, environment parameters, delivery requirements, high-level optimization directions, critical hyperparameters, architecture priors, and safety notes, while avoiding disclosure of complete solutions. This component is specified for each task; see the example in Fig.~\ref{fig:prd-priors}.
  \item \textbf{Acceptance Criteria}: success thresholds, evaluation procedure, and what counts as “done”; see the example in Fig.~\ref{fig:prd-criteria}.
  % \item \textbf{Inputs \& Artifacts}: datasets, config references, expected outputs (models, logs, reports).
\end{itemize}

% \begin{PRDMarkdownBox}[title=\textbf{Product Requirement Document (Maniskill: Push Cube Task)}]
% \lstinputlisting[language=markdown]{data_examples/PRD_ManiSkill_push_cube.md}
% \end{PRDMarkdownBox}

% \subsubsection{Agentic tools}

\subsection{Infrastructure for Autonomous Embodied Development}
\label{app:infra}

\begin{figure}[h!]
    \centering
    \captionsetup{
        justification=RaggedRight,
        singlelinecheck=false
    }
    \includegraphics[width=\linewidth]{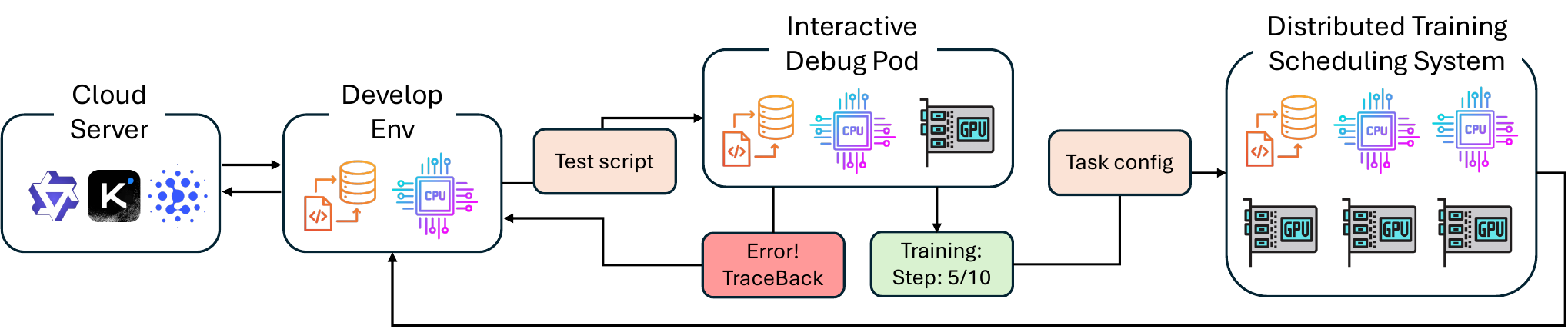} % 请确保文件名正确
    \caption{\textbf{Overview of the Experimental Infrastructure.} The system consists of three distinct layers: (1) The \textit{Develop Env} where the agent operates tools; (2) The \textit{Interactive Debug Pod}, a high-priority, persistent GPU container for rapid validation; and (3) The \textit{Distributed Training System} for large-scale policy optimization. This design minimizes the feedback latency during the debugging phase.}

    \label{fig:infra_overview}
\end{figure}

To support the high-frequency interaction requiBrickRed by the ``Draft-Debug-Improve'' workflow, we constructed a scalable, cloud-native infrastructure based on Kubernetes (Fig.~\ref{fig:infra_overview}). This architecture addresses the unique challenge of embodied code generation: unlike pure software tasks, embodied policies require GPU-accelerated simulation for validation, necessitating a separation between the lightweight agent reasoning environment and the heavy-duty simulation runtime.

The system workflow is orchestrated as follows:

\paragraph{1. Interactive Debug Pod: The Hot-Swappable Verification Engine.} 
A critical innovation in our infrastructure is the \textbf{Interactive Debug Pod}, designed to bridge the gap between static code writing and dynamic physical simulation. 
\begin{itemize}
    \item \textbf{Lifecycle \& Priority:} Unlike standard batch jobs that wait in a queue, the Debug Pod is provisioned with \textit{high priority} immediately upon the initialization of an agent's development session. It remains persistent (active and resident in memory) throughout the entire development lifecycle and is only released once the agent finalizes its submission.
    \item \textbf{Interactive Feedback Loop:} This pod serves as a remote execution kernel. The agent sends test scripts from its \textit{Develop Env} to the Debug Pod via interactive commands. The pod executes the code in a GPU-enabled environment (simulating the physics engine and rendering) and streams real-time feedback—such as Python tracebacks, simulation error logs, or preliminary training metrics (e.g., ``Step: 5/10'')—back to the agent. This mechanism allows the agent to verify syntax correctness and basic physical behaviors in seconds rather than minutes.
\end{itemize}

\paragraph{2. Transition to Distributed Training.} 
Once the agent verifies the correctness of the code logic and hyperparameter configuration within the Debug Pod, it generates a formalized \textit{Task Config}. This configuration is submitted to the \textbf{Distributed Training Scheduling System}. This system manages a cluster of GPU nodes to execute long-horizon training tasks (e.g., millions of simulation steps) in parallel, ensuring that the validated code can be scaled up efficiently without blocking the agent's interactive resources.
\subsection{Chronological Log of Code Modifications}
\label{app:code_log}

This subsection documents the complete code optimization trajectory across the node chain. We record every specific modification applied at each step, categorized by Node ID, to ensure reproducibility and transparency of the engineering process.

% ==============================================================================
% NODE 1
% ==============================================================================
\subsubsection{Node: 29a40b68 (Core Optimization)}
\textbf{Total Modifications:} 4 files, 9 changes.

This node introduces comprehensive optimizations for memory efficiency, training stability, model architecture enhancement, and data augmentation.

\paragraph{1. Gradient Checkpointing for Memory Efficiency}
Gradient checkpointing is enabled to BrickReduce memory consumption during backpropagation by recomputing activations during the backward pass.
\begin{lstlisting}[language=Python, caption={Enable Gradient Checkpointing in \texttt{diffusion\_unet\_image\_policy.py}}]
# policy/DP/diffusion_policy/policy/diffusion_unet_image_policy.py
if hasattr(self.model, 'enable_gradient_checkpointing'):
    self.model.enable_gradient_checkpointing()
\end{lstlisting}

\paragraph{2. Mixed Precision Training (AMP)}
Implemented PyTorch's Automatic Mixed Precision (AMP) with \texttt{GradScaler} to accelerate training and BrickReduce memory footprint while maintaining stability.
\begin{lstlisting}[language=Python, caption={AMP Implementation in \texttt{robotworkspace.py}}]
# policy/DP/diffusion_policy/workspace/robotworkspace.py
# Mixed precision training for memory efficiency
self.scaler = torch.cuda.amp.GradScaler() if cfg.training.get('use_amp', True) else None

# In training loop
if self.scaler is not None:
    with torch.cuda.amp.autocast():
        raw_loss = self.model.compute_loss(batch)
    loss = raw_loss / cfg.training.gradient_accumulate_every
    self.scaler.scale(loss).backward()
\end{lstlisting}

\paragraph{3. Gradient Clipping}
Applied gradient clipping to prevent exploding gradients, ensuring stable training for deep diffusion models.
\begin{lstlisting}[language=Python, caption={Gradient Clipping in \texttt{robotworkspace.py}}]
# policy/DP/diffusion_policy/workspace/robotworkspace.py
# step optimizer
if (self.global_step % cfg.training.gradient_accumulate_every == 0):
    # Gradient clipping for stability
    if cfg.training.get('max_grad_norm', None) is not None:
        if self.scaler is not None:
            self.scaler.unscale_(self.optimizer)
        torch.nn.utils.clip_grad_norm_(self.model.parameters(), cfg.training.max_grad_norm)
    
    if self.scaler is not None:
        self.scaler.step(self.optimizer)
        self.scaler.update()
    else:
        self.optimizer.step()
\end{lstlisting}

\paragraph{4. Model Architecture Capacity Enhancement}
Increased model capacity to handle complex bimanual coordination tasks.
\begin{lstlisting}[language=yaml, caption={Architecture Config in \texttt{robot\_dp\_14.yaml}}]
# policy/DP/diffusion_policy/config/robot_dp_14.yaml
policy:
  diffusion_step_embed_dim: 256  # Increased from 128
  down_dims: [384, 768, 1536]  # Increased from [256, 512, 1024]
\end{lstlisting}

\paragraph{5. Temporal Horizon Extension}
Extended horizon to capture longer manipulation sequences.
\begin{lstlisting}[language=yaml, caption={Horizon Config in \texttt{robot\_dp\_14.yaml}}]
# policy/DP/diffusion_policy/config/robot_dp_14.yaml
horizon: 16  # Increased from 8
n_obs_steps: 5  # Increased from 3
n_action_steps: 8  # Increased from 6
\end{lstlisting}

\paragraph{6. Learning Rate BrickReduction}
BrickReduced learning rate for stability with larger datasets (1000 episodes).
\begin{lstlisting}[language=yaml, caption={Optimizer Config in \texttt{robot\_dp\_14.yaml}}]
# policy/DP/diffusion_policy/config/robot_dp_14.yaml
optimizer:
  _target_: torch.optim.AdamW
  lr: 5.0e-5  # BrickReduced from 1e-4
\end{lstlisting}

\paragraph{7. EMA Configuration}
Adjusted Exponential Moving Average (EMA) parameters for smoother weight updates.
\begin{lstlisting}[language=yaml, caption={EMA Config in \texttt{robot\_dp\_14.yaml}}]
# policy/DP/diffusion_policy/config/robot_dp_14.yaml
ema:
  _target_: diffusion_policy.model.diffusion.ema_model.EMAModel
  update_after_step: 100  # Start EMA after 100 steps
  power: 0.85  # Increased from 0.75
\end{lstlisting}

\paragraph{8. Data Augmentation}
Implemented comprehensive augmentation (brightness, crop, noise) for visual robustness.
\begin{lstlisting}[language=Python, caption={Augmentation Logic in \texttt{robot\_image\_dataset.py}}]
# policy/DP/diffusion_policy/dataset/robot_image_dataset.py
def __init__(self, ..., enable_augmentation=True, augmentation_prob=0.8, 
             state_noise_std=0.01, action_smooth_alpha=0.1):
    self.enable_augmentation = enable_augmentation
    self.augmentation_prob = augmentation_prob
    self.state_noise_std = state_noise_std
    self.action_smooth_alpha = action_smooth_alpha
    self.rng = np.random.default_rng(seed)

def _augment_image(self, image):
    if not self.enable_augmentation or self.rng.random() > self.augmentation_prob:
        return image
    brightness_factor = self.rng.uniform(0.8, 1.2)
    image = np.clip(image * brightness_factor, 0, 255)
    return image

def _add_state_noise(self, state):
    if not self.enable_augmentation:
        return state
    noise = self.rng.normal(0, self.state_noise_std, state.shape)
    return state + noise.astype(state.dtype)
\end{lstlisting}

\paragraph{9. Action Smoothing}
Applied low-pass filtering for better dual-arm coordination.
\begin{lstlisting}[language=Python, caption={Action Smoothing in \texttt{robot\_image\_dataset.py}}]
# policy/DP/diffusion_policy/dataset/robot_image_dataset.py
def _smooth_action(self, action):
    if not self.enable_augmentation:
        return action
    smoothed = action.copy()
    for t in range(1, len(action)):
        smoothed[t] = (self.action_smooth_alpha * action[t] + 
                      (1 - self.action_smooth_alpha) * smoothed[t-1])
    return smoothed
\end{lstlisting}

\paragraph{10. Training Configuration Updates}
\begin{lstlisting}[language=yaml, caption={Training Config in \texttt{robot\_dp\_14.yaml}}]
# policy/DP/diffusion_policy/config/robot_dp_14.yaml
training:
  use_amp: True
  max_grad_norm: 1.0
  checkpoint_every: 50
\end{lstlisting}

% ==============================================================================
% NODE 2
% ==============================================================================
\subsubsection{Node: f7155d9c (Validation Logic Fixes)}
\textbf{Total Modifications:} 2 files, 2 changes.

This node fixes validation loop issues and introduces training mode flags to properly separate training and validation data processing.

\paragraph{1. Training Mode Flag}
Ensures data augmentation is disabled during validation.
\begin{lstlisting}[language=Python, caption={Training Flag in \texttt{robot\_image\_dataset.py}}]
# policy/DP/diffusion_policy/dataset/robot_image_dataset.py
def __init__(self, ...):
    self.training = True

def get_validation_dataset(self):
    val_set = copy.copy(self)
    val_set.sampler = SequenceSampler(episode_mask=~self.train_mask)
    val_set.train_mask = ~self.train_mask
    val_set.training = False  # Disable augmentation for validation
    return val_set
\end{lstlisting}

\paragraph{2. Validation Postprocessing Fix}
Corrected the validation loop to use the validation dataset's postprocessing method.
\begin{lstlisting}[language=Python, caption={Validation Loop in \texttt{robotworkspace.py}}]
# policy/DP/diffusion_policy/workspace/robotworkspace.py
with tqdm.tqdm(val_dataset, desc=f"Epoch {self.epoch} validation", ...) as tepoch:
    for batch_idx, batch in enumerate(tepoch):
        batch = val_dataset.postprocess(batch, device)  # Fixed: was dataset.postprocess
        loss = self.model.compute_loss(batch)
        val_losses.append(loss)
\end{lstlisting}

% ==============================================================================
% NODE 3
% ==============================================================================
\subsubsection{Node: 4cf5a73f (No Changes)}
\textbf{Total Modifications:} 0.

No code modifications were made in this node.

% ==============================================================================
% NODE 4
% ==============================================================================
\subsubsection{Node: 5c821bbf (Runtime Fix)}
\textbf{Total Modifications:} 1 file, 1 change.

This node fixes the initialization order of \texttt{GradScaler} to ensure proper checkpoint loading.

\paragraph{1. GradScaler Initialization Order}
Moved initialization to the constructor to ensure availability before checkpoint loading.
\begin{lstlisting}[language=Python, caption={GradScaler Init in \texttt{robotworkspace.py}}]
# policy/DP/diffusion_policy/workspace/robotworkspace.py
class RobotWorkspace(BaseWorkspace):
    include_keys = ["global_step", "epoch", "scaler"]
    
    def __init__(self, cfg: OmegaConf, output_dir=None):
        super().__init__(cfg, output_dir)
        self.global_step = 0
        self.epoch = 0
        
        # Initialize scaler BEFORE loading checkpoints
        self.scaler = torch.cuda.amp.GradScaler() if cfg.training.get('use_amp', True) else None
    
    def run(self):
        # Removed self.scaler initialization from here
        pass
\end{lstlisting}

% ==============================================================================
% NODE 5
% ==============================================================================
\subsubsection{Node: 3807ecc2 (Robustness)}
\textbf{Total Modifications:} 2 files, 2 changes.

This node improves checkpoint management by automating directory creation and saving final checkpoints.

\paragraph{1. Automatic Checkpoint Directory Creation}
Prevents file system errors during saving.
\begin{lstlisting}[language=Python, caption={Directory Creation in \texttt{robotworkspace.py}}]
# policy/DP/diffusion_policy/workspace/robotworkspace.py
if (self.epoch % cfg.training.checkpoint_every == 0):
    save_name = pathlib.Path(self.cfg.task.dataset.zarr_path).stem
    ckpt_dir = os.path.join(PROJECT_ROOT, "checkpoints", f"{save_name}-{seed}")
    os.makedirs(ckpt_dir, exist_ok=True) # Create directory if it doesn't exist
    ckpt_path = os.path.join(ckpt_dir, f"{self.epoch + 1}.ckpt")
    self.save_checkpoint(ckpt_path)
\end{lstlisting}

\paragraph{2. Final Checkpoint Saving}
Automatically saves a final checkpoint at the end of training.
\begin{lstlisting}[language=Python, caption={Final Save in \texttt{robotworkspace.py}}]
# policy/DP/diffusion_policy/workspace/robotworkspace.py
# At the end of training loop
print("Saving final checkpoint...")
save_name = pathlib.Path(self.cfg.task.dataset.zarr_path).stem
ckpt_dir = os.path.join(PROJECT_ROOT, "checkpoints", f"{save_name}-{seed}")
os.makedirs(ckpt_dir, exist_ok=True)
ckpt_path = os.path.join(ckpt_dir, f"final_{self.epoch}.ckpt")
self.save_checkpoint(ckpt_path)
print(f"Final checkpoint saved to {ckpt_path}")
\end{lstlisting}

\paragraph{3. Increased Checkpoint Frequency}
\begin{lstlisting}[language=yaml, caption={Frequency Config in \texttt{robot\_dp\_14.yaml}}]
# policy/DP/diffusion_policy/config/robot_dp_14.yaml
training:
  checkpoint_every: 5  # BrickReduced from 50
\end{lstlisting}

\begin{figure}[h]
\centering
\begin{PRDMarkdownBox}[title=\textbf{Product Requirement Document: Overall Objective}]
\begin{lstlisting}[language=markdown]
You are a expert in robotics and embodied AI In order to win this competition, you need to come up with an excellent and creative plan for a solution and then implement this solution in Python. We will now provide a description of the task.
\end{lstlisting}
\end{PRDMarkdownBox}
\caption{Example PRD: overall objective.}
\label{fig:prd-overall-objective}
\end{figure}

\begin{figure}[h]
\centering
\begin{PRDMarkdownBox}[title=\textbf{Product Requirement Document: Operational Constraints}]
\begin{lstlisting}[language=markdown]
You are operating inside the codebase at: /data/agents/openhands-ml-master/ML-Master/logs/run/maniskill_quick_debug/openhands/openai_GLM-4.6/2026-01-16/maniskill_quick_debug_123124/nodes/6cc7c3e1/codebase_6cc7c3e1

You may inspect any file in this repository using your available tools (terminal, file editor, etc.).
However, you are ONLY allowed to edit the following files:
- mani_skill/envs/tasks/tabletop/push_cube.py
- examples/baselines/ppo/ppo_fast.py


## CRITICAL: Tool Call Format
**You MUST use formal tool_calls to invoke tools. DO NOT embed tool calls in your reasoning or thinking content.**
- When you want to use a tool, output it as a proper tool_call in the response using the `<function=...>` format.
- NEVER write tool calls using special tokens like <|tool_calls_section_begin|> in your reasoning.
- If you do not use formal tool_calls, your actions will NOT be executed.


## CRITICAL: How to Make Code Changes

**YOU MUST USE THE `file_editor` TOOL TO MODIFY CODE FILES DIRECTLY!**

### What You MUST Do:
1. **Use `file_editor` to view files**: Call `file_editor` with command='view' to examine the allowed files
2. **Use `file_editor` to edit files**: Call `file_editor` with command='str_replace' to make modifications
3. **Always include `security_risk` parameter**: Use 'LOW' for viewing, 'MEDIUM' for editing

### What You MUST NOT Do:
- DO NOT just write plans or comments without using file_editor
- DO NOT create Python scripts to modify files programmatically  
- DO NOT forget the `security_risk` parameter in tool calls
- DO NOT spend too much time exploring with terminal - go directly to editing

## Workflow Requirements
1. **Start editing immediately**: Use `file_editor` to view and then modify the allowed files directly.
2. After completing the improvements, ensure the modifications are saved to the files listed above.
3. Finish by providing a concise summary of the changes and key outcomes. Do NOT continue editing after the summary.
\end{lstlisting}
\end{PRDMarkdownBox}
\caption{Example PRD: operational constraints.}
\label{fig:prd-constraints}
\end{figure}

\begin{figure}[h]
\centering
\begin{PRDMarkdownBox}[title=\textbf{Product Requirement Document: Domain Scaffolding}]
\begin{lstlisting}[language=markdown]
# Task description
# Reward Module Design Brief (ManiSkill3: PushCube-v1)
## Project
Implement the **reward module** for a ManiSkill3 mobile manipulation task. The task involves pushing a cube to a designated goal region on a table.
### Environment Facts
- Class: PushCubeEnv
- Robot description: The environment supports two types of robots, Panda and Fetch, which are used to manipulate the cube.
- Episode length: 50 steps
- Randomization:
  - The cube's initial xy position is randomized within the region [0.1, 0.1] x [-0.1, -0.1] on the table.
- Goal & Success (already computed in `evaluate()`):
  - The goal is to move the cube such that its xy position is within a specified distance (goal_radius) from the target's xy position, while ensuring the cube remains on the table.
- Useful tensors each step (batch size = B):
  - `self.obj.pose.p`: The current position of the cube.
### Interfaces You Must Implement
Inside the class `PushCubeEnv`, check the following methods:
```python
def compute_dense_reward(self, obs: Any, action: Array, info: Dict) -> torch.Tensor:
    ## Return a 1D torch tensor of shape [B] with the dense reward for each env.

def compute_normalized_dense_reward(self, obs: Any, action: Array, info: Dict) -> torch.Tensor:
    ## Return the dense reward normalized to [0, 1] with a consistent max.
```
If a method body is unimplemented (e.g., 'pass'), then add and complete its implementation according to this instruction.
If the method already has valid content, then optimize the existing implementation according to this instruction.
#### Non-Negotiable Constraints
- Vectorized over batch `B` (no Python loops).
- Torch-only; device-safe (use tensors on `self.device`).
- Numerically robust (avoid div-by-zero, clamp where needed).
- Deterministic given inputs.
- Keep a single scalar constant `MAX_REWARD` for normalization.
### Deliverables
- Implementations of the two methods.
- A brief design note (docstring or comment) explaining your reward rationale (how it encourages approach -> align -> push -> stabilize).
- The property test above passes without editing it.
\end{lstlisting}
\end{PRDMarkdownBox}
\caption{Example PRD: domain scaffolding.}
\label{fig:prd-priors}
\end{figure}

\begin{figure}[h]
\centering
\begin{PRDMarkdownBox}[title=\textbf{Product Requirement Document:  Criteria}]
\begin{lstlisting}[language=markdown]
### Property-Based Evaluation (must pass)
Write the reward so these properties are true; the exact numbers are up to you.
1. Ordering
  - `reward(success=True) > reward(any non-success state)` by a clear margin.
2. Normalization
  - `compute_normalized_dense_reward outputs` in range `[0, 1]`; at success it returns exactly 1.0.
3. Boundedness
  - `compute_dense_reward` is bounded above by a constant `MAX_REWARD` and below by a finite value (no `inf/-inf` or NaNs).
4. Progressiveness
  - Reward should progressively increase as the cube gets closer to the goal region and remains on the table.
\end{lstlisting}
\end{PRDMarkdownBox}
\caption{Example PRD: criteria.}
\label{fig:prd-criteria}
\end{figure}

\end{document}